\documentclass[manuscript]{acmart} 

\usepackage{pdfpages}
\usepackage{array}
\usepackage{tabu}
\usepackage{hhline}
\usepackage{multirow}

\AtBeginDocument{%
  \providecommand\BibTeX{{%
    \normalfont B\kern-0.5em{\scshape i\kern-0.25em b}\kern-0.8em\TeX}}}

\setcopyright{acmcopyright}
\copyrightyear{2022}
\acmYear{2022}

\begin{document}






\title{Neuromophic High-Frequency 3D Dancing Pose Estimation in Dynamic Environment}

\author{Zhongyang Zhang}
\email{zhz138@ucsd.edu}
\affiliation{%
  \institution{University of California San Diego}
  \city{San Diego}
  \state{California}
  \country{USA}
}

\author{Kaidong Chai}
\email{kchai@umass.edu}
\affiliation{
  \institution{University of Massachusetts Amherst}
  \city{Amherst}
  \state{Massachusetts}
  \country{USA}}

\author{Haowen Yu}
\email{hyu@cs.umass.edu}
\affiliation{
  \institution{University of Massachusetts Amherst}
  \city{Amherst}
  \state{Massachusetts}
  \country{USA}}

\author{Ramzi Majaj}
\email{majaj@umass.edu}
\affiliation{
  \institution{University of Massachusetts Amherst}
  \city{Amherst}
  \state{Massachusetts}
  \country{USA}}

\author{Francesca Walsh}
\email{fnwalsh@umass.edu}
\affiliation{
  \institution{University of Massachusetts Amherst}
  \city{Amherst}
  \state{Massachusetts}
  \country{USA}}

\author{Edward Wang}
\email{ejaywang@eng.ucsd.edu}
\affiliation{
  \institution{University of California San Diego}
  \city{San Diego}
  \state{California}
  \country{USA}}

\author{Upal Mahbub}
\email{upalmahbub@yahoo.com}
\affiliation{
  \institution{Independent Scholar}
  \city{San Diego}
  \state{California}
  \country{USA}}

\author{Hava Siegelmann}
\email{hava@umass.edu}
\affiliation{
  \institution{University of Massachusetts Amherst}
  \city{Amherst}
  \state{Massachusetts}
  \country{USA}}

\author{Donghyun Kim}
\email{donghyunkim@cs.umass.edu}
\affiliation{
  \institution{University of Massachusetts Amherst}
  \city{Amherst}
  \state{Massachusetts}
  \country{USA}}

\author{Tauhidur Rahman}
\email{trahman@ucsd.edu }
\affiliation{%
  \institution{University of California San Diego}
  \city{San Diego}
  \state{California}
  \country{USA}}

\renewcommand{\shortauthors}{Zhongyang Zhang and Tauhidur Rahman, et al.}

\begin{abstract}

As a beloved sport worldwide, dancing is getting integrated into traditional and virtual reality-based gaming platforms nowadays. It opens up new opportunities in the technology-mediated dancing space. These platforms primarily rely on passive and continuous human pose estimation as an input capture mechanism. Existing solutions are mainly based on RGB or RGB-Depth cameras for dance games. The former suffers in low-lighting conditions due to the motion blur and low sensitivity, while the latter is too power-hungry, has a low frame rate, and has limited working distance. With ultra-low latency, energy efficiency, and wide dynamic range characteristics, the event camera is a promising solution to overcome these shortcomings. We propose YeLan, an event camera-based 3-dimensional high-frequency human pose estimation(HPE) system that survives low-lighting conditions and dynamic backgrounds. We collected the world's first event camera dance dataset and developed a fully customizable motion-to-event physics-aware simulator. YeLan outperforms the baseline models in these challenging conditions and demonstrated robustness against different types of clothing, background motion, viewing angle, occlusion, and lighting fluctuations.

\end{abstract}


\maketitle

\section{Introduction}

Dancing is a popular activity, loved by people all over the world. In recent years, Technology Mediated Dancing (TMD) is gaining popularity as it supports remote playful and healthy physical activities in the form of dancing \cite{lopez2013efectividad, cheng2017effects, marquez2017regular}. From remote control-based gaming console games to Virtual Reality (VR) platforms, different forms of TMD are finding their way into users' living spaces. For all of these TMDs, human pose estimation (HPE) plays a critical role as it infers the user's unique, complex, and fast-changing dance poses to interact with the computer. To adapt to as many players as possible, TMD requires high-fidelity HPE that works robustly in various challenging and realistic indoor environments (e.g., in dynamic lighting and background conditions).

The state of the art HPE systems are generally developed based on depth and RGB cameras. However, both the standard RGB camera-based and depth-based monocular HPE systems \cite{Chen2022AnatomyAware3H, Chen2020MonocularHP, Hassan2019Resolving3H} fails to generate ultra-fast/high-speed human pose inferences with its limited frame rate which is a critical requirement for many real world applications (e.g., virtual reality dance game, high frequency motion characterization for tremor monitoring applications). In one hand RGB based HPE fails to operate in low-lighting conditions due to the severe motion blur-related issues, on the other hand the depth camera can only operate upto a limited distance and often fails with darker clothing and skin tone. Additionally, both cameras do not inherently differentiate between static and moving objects. Consequently, both static and dynamic (or moving) objects in the background are captured as well as the target human body which impacts the robustness of HPE in dynamic settings. The depth camera uses an active infrared light and consequently has high power requirement. In this proposed work, we aim to develop a high-speed/low latency and low power 3D human pose estimation framework that can operate in dynamic lighting and motion situations.


Event camera \cite{lichtsteiner2008128}, also known as Dynamic Vision Sensor(DVS) or neuromorphic camera, is a silicon retina design based on one of the core mechanisms of mammalian vision that makes it particularly sensitive to moving targets and changing lighting conditions. Event cameras are based on the premise that for mammals, moving objects often contain more valuable information for hunting and escaping from predators, while relatively static backgrounds do not deserve much attention for constant monitoring and processing. By imitating this dynamic vision characteristic, each pixel of an event camera works asynchronously and keeps track of extremely subtle brightness changes independently in log-scale, allowing event cameras to be equally sensitive to motion in both high and low-lighting conditions. This mechanism automatically filters out the static background and does not need to wait to transmit the entire, detailed but bulky frame every time something happens. Alternatively, moving targets can rapidly trigger a significant number of events with super-high time resolution, making event cameras more sensitive to dynamic targets. The wide dynamic range makes event cameras robust to various situations, from night to glaring noon, from the tunnel to night driving. Beyond advantages with low-light conditions, event cameras also are less affected by skin color, and brightness change \cite{4541871, gallego2020event}.

DVS HPE has attracted much attention recently due to these advantages described, and in response, some significant datasets have been collected \cite{Calabrese2019DHP19DV, scarpellini2021lifting}. Besides the fact that these datasets use fixed patterns for motion guidance instead of dances, they also have significant issues around utility for developing practical systems. The data collection in these studies is conducted in ideal lab environments, with no background moving contents to serve as interference. The lighting is adequate and stable, limiting them to address low-and/or-changing lighting conditions. These uncertainties and noise reflect the real world, and
introduce realistic challenges that must be answered to make this technology more practical and generalizable. We introduce two newly collected datasets. One is a real-world dataset with dynamic background under high and low-lighting conditions. The other is a simulated dataset with a fully controllable and customizable pipeline to generate new data pieces on demand. These datasets serve this work requirement perfectly, and will also be beneficial to the community as they will be made publicly accessible. The comparison between existing event camera-based HPE dataset and our collected datasets are listed below in Table. \ref{tab:datasets}.

\begin{table}[]
\caption{The comparison between existing 3D HPE datasets and our collected datasets.}
\begin{tabular}{|l|l|l|l|ll|}
\hline
Study          & Human3.6M\cite{ionescu2013human3}  & MKV\cite{zimmermann20183d}         & DHP19\cite{Calabrese2019DHP19DV}                                                                    & \multicolumn{2}{l|}{YeLan}                                                                     \\ \hline
Modality       & RGB Camera & RGBD Camera & Event Camera                                                             & \multicolumn{2}{l|}{Event Camera}                                                              \\ \hline
Inference Rate & 50 FPS     & 10 FPS      & \begin{tabular}[c]{@{}l@{}}Arbitrary,\\ Tested up to 50 FPS\end{tabular} & \multicolumn{2}{l|}{\begin{tabular}[c]{@{}l@{}}Arbitrary,\\ Tested up to 150 FPS\end{tabular}} \\ \hline
Lighting       & High       & High        & High                                                                     & \multicolumn{2}{l|}{Low to High}                                                               \\ \hline
Data Type      & Real World & Real World  & Real World                                                               & \multicolumn{1}{l|}{Synthetic}                       & Real World                              \\ \hline
Background     & Static     & Static      & Static                                                                   & \multicolumn{1}{l|}{Dynamic}                         & Static \& Dynamic                       \\ \hline
View           & Arbitrary  & Arbitrary   & Arbitrary                                                                & \multicolumn{1}{l|}{Arbitrary}                       & Front                                   \\ \hline
Clothing       & Arbitrary  & Arbitrary   & Tight                                                                    & \multicolumn{1}{l|}{Arbitrary}                       & Tight                                   \\ \hline
Data Size      & 3,600,000  & 22406       & 350,860                                                                  & \multicolumn{1}{l|}{3,958,169}                       & 446,158                                 \\ \hline
\end{tabular}
\label{tab:datasets}
\end{table}

Beyond datasets being too idealistic, previous DVS HPE works are also limited by the missing torso problem, which is fundamental to the dynamic characteristics of the event camera. When some parts of the human body stay still, event cameras will only capture other moving parts of the body and ignore these static parts. Therefore, the event representation during this period contains little or no information about them, resulting in significant estimation errors.



To build a DVS HPE system that could work in more realistic environments, we proposed a two-stage system \emph{YeLan} that accurately estimates human poses under low-lighting conditions with noisy background contents. The name \emph{YeLan} is derived from the character name in an anime game, Genshin Impact, which literally means ``night orchid''. Interestingly, it can also be interpreted as another Chinese word ``night viewing''. An early-exit-style mask prediction network is implemented in the first stage to help remove the moving background objects while saving as much energy as possible. BiConvLSTM is applied in the second stage to enable the flow of information between frames, which helps with the missing torso problem. Also, TORE volume is adopted to build denser and more informative input tensors and help solve the low event rate problem in low-light conditions. We conducted massive experiments to compare our method with baseline DVS HPE methods and achieved SOTA on the proposed two new datasets. 

In summary, the core contribution of the paper are as follows.

\begin{enumerate}
    \item We proposed YeLan, which to the best of our knowledge, is the first event camera-based 3D human pose estimation solution specially for dance. It works robustly under various challenging conditions, including low-lighting and occlusion. YeLan overcame the inherent disadvantages of the event camera and fully exploited its strengths.
    \item We built an end-to-end simulator that enables precise and low-level control of the generated events. The simulator is garment physics-aware, highly customizable, and extensible.
    \item As far as we know, the generated synthetic dataset from our simulator is the first event camera dataset designed for dance, and is also the largest event camera HPE dataset till now. It has unprecedented variability compared to existing datasets.
    \item We conducted a human subject study and collected a real-world dance HPE dataset with low-lighting conditions and mobile background content considered.
    \item An early-exit-style network is adopted in our mask prediction module, which boosts efficiency significantly.
    \item We validated YeLan on both datasets, and it outperform all baseline models. 
\end{enumerate}



\section{Related Work}

\textbf{Dance and Its Effect:} From teenagers to elders, dance is a beloved communication and sport modality by people worldwide with a long history. Due to the high practicality and the cultural difference, dance develops a wide range of variations in different regions and times containing diverse styles, rhythms, intensities, and steps. Dance is proven to have a prominent positive influence on physical and mental health, with many solid studies as evidence \cite{sanchez2021characterization}. A recent study \cite{teixeira2019dance} demonstrated that dance could positively impact neuroplasticity and enhance neural activation in several brain regions. As a result, dance can be used as a rehabilitation tool for various brain-related pathologies \cite{sanchez2021characterization}. Recent literature has documented that dance as therapy has a significant positive effect on depression \cite{akandere2011effect}, schizophrenia \cite{cheng2017effects}, Parkinson's \cite{hashimoto2015effects}, fibromyalgia \cite{lopez2013efectividad, del2012comparacion}, dementia \cite{borges2018effects}, cognitive deterioration \cite{marquez2017regular, zhu2018effects}, stress \cite{pinniger2012argentine} and chronic stroke \cite{patterson2018dance}. 


Digital and mobile technology assist dancers in all aspects. Technology-assisted dance is increasingly becoming popular in recent years \cite{hsueh2019understanding}. It makes the dance more accessible and easier to learn \cite{ruth2020exergames,romeroeffectiveness}. Dance has been combined with different video games \cite{hsueh2019understanding,kloos2013video,ruth2020exergames,adcock2020usability}, including \textit{Just Dance Series} \cite{JustDanceVideo2022}, \textit{Dance Dance Revolution} \cite{DanceDanceRevolution2022}, and \textit{Dance Central} \cite{DanceCentral2022}. Moreover, there are some other movement-based VR rhythm games like \textit{Beat Saber} \cite{BeatSaber2022}, \textit{Synth Riders} \cite{SynthRidersFreestyleDance}, and \textit{Dance Collider} \cite{DANCECOLLIDER}. Technology-assisted dance also includes video-based remote coaching, Virtual Reality (VR) or Augmented Reality (AR)-based dance games. 

High fidelity 3D human pose estimation is a crucial component of Technology-Assisted Dance as they can turn dance gestures as input or commands \cite{Alaoui2013DanceIW, Alaoui2012MovementQA} to the AR/VR or traditional video-based dance games. The 3D human pose estimates assist in performance evaluation, personalized feedback, and choreography \cite{hsueh2019understanding}. However, traditional RGB and depth camera-based 3D human pose estimates often fail to capture the rapid/high-speed changes in poses during a dance performance in challenging conditions (e.g., low light, dynamic background). In this work, we introduce an Event camera-based 3D human pose estimation to support technology-mediated dance in these challenging and dynamic conditions.



\textbf{Human Pose Estimation:} Depending on taxonomy, existing methods for 3D human pose estimation can be classified by modality and the number of used sensors. If classified by modality, the majority of approaches are RGB-based \cite{Hassan2019Resolving3H, Chen2020MonocularHP, Omran2018NeuralBF} or RGB-Depth-based \cite{rim2020real, srivastav2018mvor, michel2017markerless, zimmermann20183d}. Generally speaking, RGB image-based methods have fewer requirements on equipment and are more comprehensively explored. On the other hand, the RGB-Depth camera gains an advantage by introducing additional depth information, which is beneficial to detection, segmentation, and parts localization. However, depth cameras rely heavily on IR projectors to build the depth map, which is highly power-hungry and vulnerable to bright environmental light, limiting the RGBD cameras' working distance and FOV, making them hard to be applied on the outside. For the second classification method, the resulting two categories are: monocular \cite{Chen2020MonocularHP, zimmermann20183d} and multi-view \cite{Rhodin2018LearningM3, srivastav2018mvor} methods. The multi-view method observes subjects from multiple views/cameras from different vantage points \cite{Ge2016Robust3H}. The multi-view method requires significantly more power. More importantly, setting up a multi-view event camera data collection system is complex and expensive, making applications outside the lab very difficult. The monocular method estimates human pose from a single camera view. Methods for human pose estimation methods can also be classified in another way: model-based \cite{Omran2018NeuralBF}, and skeleton-based \cite{Chen2022AnatomyAware3H} methods. While the model-based method attempts to reconstruct the full 3D body shape of a human model \cite{Hassan2019Resolving3H}, skeleton-based methods use a bone skeleton as an intermediate representation and regress the joint locations in 3D space. In this work, we use a monocular skeleton-based approach that can allow us to achieve a more efficient and practical solution for 3D human pose estimation in real-world settings.

\textbf{Dynamic Vision Sensor(DVS)}, or event camera was originally proposed by Lichtsteiner et al. \cite{lichtsteiner2008128}. In recent years, event camera has gained more attention increasingly. It has been applied to many computer vision tasks, including object recognition \cite{li2021graph, kim2021n}, segmentation \cite{alonso2019ev}, corner detection \cite{yilmaz2021evaluation, mohamed2021dynamic}, gesture recognition \cite{Calabrese2019DHP19DV, Wang2019EVGaitER, Ghosh2019SpatiotemporalFF}, optical flow estimation \cite{brebion2021real, liu2022edflow}, depth estimation \cite{gehrig2021combining, hidalgo2020learning}, Simultaneous Localization And Mapping (SLAM) \cite{jiao2021comparing, bertrand2020embedded}, and autonomous driving \cite{li2019event, chen2020event}. While RGB cameras struggle due to motion blur, event camera, by design, is highly sensitive to lux variation in both extremely overexposed and underexposed scene \cite{Berthelon2018EffectsOC}. 

While the event camera has already been proposed for human pose estimation in existing literature, the prior works have primarily focused on designing algorithms that are relatively noise-free, background-activity-less, and well-lit settings. For example, a recent event camera HPE dataset, DHP19 \cite{Calabrese2019DHP19DV} contains a series of event recordings of human movements, poses, and moving objects. Some event camera datasets are used in gesture recognition \cite{Amir2017ALP}, and action recognition \cite{Hu2016DVSBD}. However, these datasets capture a limited number of motion trajectories in noise-free controlled environments. The lack of a diversified dataset in realistic conditions under low and dynamic light is a major bottleneck for developing a robust event camera-based sensing system. In this work, we will demonstrate how an Event Camera-based mobile sensing platform can effectively capture high-frequency 3D human poses during a dance performance while being highly resilient to different real-world challenging conditions, including low light, dynamic moving background, higher sensor field of view, longer distance between the sensor and target human, and diverse outfit or clothing.

\section{Design Consideration}
People can dance or play games in various environments, which is not controllable by developers. Dancers or gamers may dance in a low light indoor environment or be disturbed by surrounding moving objects. Estimating human pose in low-lighting conditions in the presence of surrounding activities with an event camera brings many new challenges. While low-lighting condition results in fewer events for the same movement and lowers the signal-noise ratio of the event representation, background activities and motion from surrounding objects introduce additional events. These events baffle the human pose estimation model to differentiate the target human body motion-related events from the background activities. In some cases, events triggered by background objects' movements are even an order of magnitude larger than by the foreground subject, especially in low-lighting conditions. YeLan proposes to solve this challenge with an event filtering mechanism that predicts binary-mask over the target human body and rejects all events that are not generated due to the user's body movement.




One of the significant strengths of the event camera is that each pixel can asynchronously respond to a tiny amount of light intensity change and trigger events with minimal latency. Consequently, the entire imaging grid can produce a massive event sequence from a single camera. One fundamental challenge is learning a representation that will preserve meaningful spatio-temporal information about the human pose. Another major challenge for event camera-based human pose estimation is the ``missing torso problem''. Even during physical activity, we do not move our body parts equally. For example, the upper body might be in motion while the lower body is stationary. This partial immobility results in the silence of corresponding pixels within the event camera, which may further result in higher error in predicting these missing joints.

Prior work on event camera proposed different representations, including event frame \cite{rebecq2017real}, constant-count frame \cite{maqueda2018event}, time surface \cite{benosman2013event}, and voxel grid \cite{zhu2019unsupervised}. Time surface is a 2D representation where each pixel records the most recent event's timestamp as its pixel value. Each pixel stores the timestamp of the last event \cite{Lagorce2017HOTSAH} in that location. A voxel grid is a 3D histogram of events. Each voxel represents the number of events in an interval at a pixel location. Voxel grid prevents information loss by preserving spatial-temporal information of the whole history instead of collapsing the history into a 2D grid representation. However, event frame, event count, or voxel grid do not preserve relatively distant historical information, which gives rise to the missing torso problem. Time surface, on the other hand, can discard the temporal information and cannot keep the information from multiple events at the same pixel.
Moreover, these existing representations suffer from a low signal-to-noise ratio (SNR) in low-lighting conditions. In this work, we have used a modified Time Ordered Recent Event (TORE) volume representation \cite{Baldwin2021TimeOrderedRE} that can simultaneously preserve both the latest and short historical information, which can then compensate for the missing torso problem. It also serves as a noise filter for noises like salt and paper without affecting other informative signals.



\section{Proposed System}\label{pipeline}
%


\subsection{Event Preprocessing and TORE Volume}

\begin{figure*}[t]
    \centering
    \includegraphics[width=0.8\linewidth]{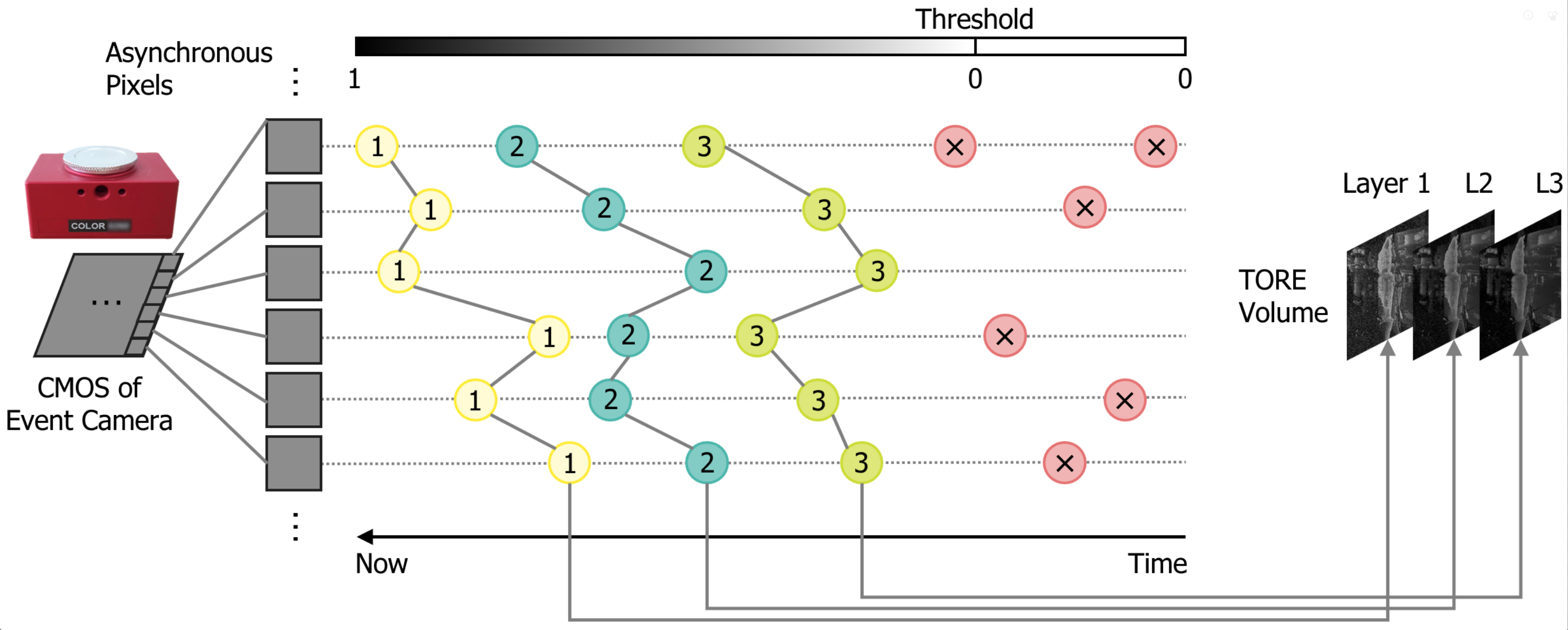}
    \caption{The conceptual figure of the modified TORE volume.}
    \label{fig:tore_concept}
\end{figure*}

TORE \cite{Baldwin2021TimeOrderedRE} attempts to mimic the human retina by preserving the membrane's potential properties. A fixed-length(e.g., $K$) First-In-First-Out (FIFO) queue is adopted to record the relative timestamp of the most recent $K$ events. When a new event comes into a pixel's queue, its relative timestamp is pushed in, and the oldest event in the queue gets popped out. TORE computes the logarithm of these timestamps in the FIFO buffer. TORE transforms the sparse events stream into a dense and bio-inspired representation with minimum information loss and achieves SOTA in many DVS tasks(e.g., classification, denoising, human pose estimation). An exhaustive comparison of different representations could be found in \cite{Baldwin2021TimeOrderedRE}.

We use a modified TORE volume representation: Normalization, 0-1 Flip, and a range scaling. Firstly, values inside the TORE volume are normalized to the range [0,1] for faster convergence. Then we flipped the normalized maximum and minimum values. By default, the oldest events and pixels with no event recorded are given the maximum value $log(\tau)$, which is counter-intuitive and may harm the convergence speed. We can easily solve this problem by letting the value $v$ in TORE be set to $log(\tau)-v$. With this modification, older events have a smaller value while newer events have a greater value, and the pixel where no event is recorded is set to 0. Lastly, because logarithm is applied in the TORE calculation, the most recent 4.63$\mu$s takes the $[0.7,1]$ range of $[0,1]$, and 4.63$\mu$s is smaller than the DVS camera's sensitivity time (150ms is used in the original TORE paper \cite{Baldwin2021TimeOrderedRE}). So here is our modification formula:
\begin{align}
    v'=(1-v/log(\tau))|_{[0,0.7]}/0.7
\end{align}
where $v$ and $v'$ represents the original and modified TORE value, while the notation $x|_[a,b]$ means the value of $x$ is clamped within the range $[a,b]$.

TORE holds a first-in-first-out queue separately for each pixel, and the value in the queue drops as time goes on. TORE can keep the history of pixels up to five seconds. These characteristics make it suitable to work in dynamic conditions. TORE can drop redundant histories on the same pixel when in high-lighting conditions, relieving the model's computational pressure. When in low-lighting conditions, the event histories captured in the past could compensate and help decide the position of some joints that do not move significantly during the last time window and thus help solve the missing torso problem. Fig. \ref{fig:tore_concept} shows the details to generate the modified TORE volumes.

\begin{figure*}
    \centering
    \includegraphics[width=\linewidth]{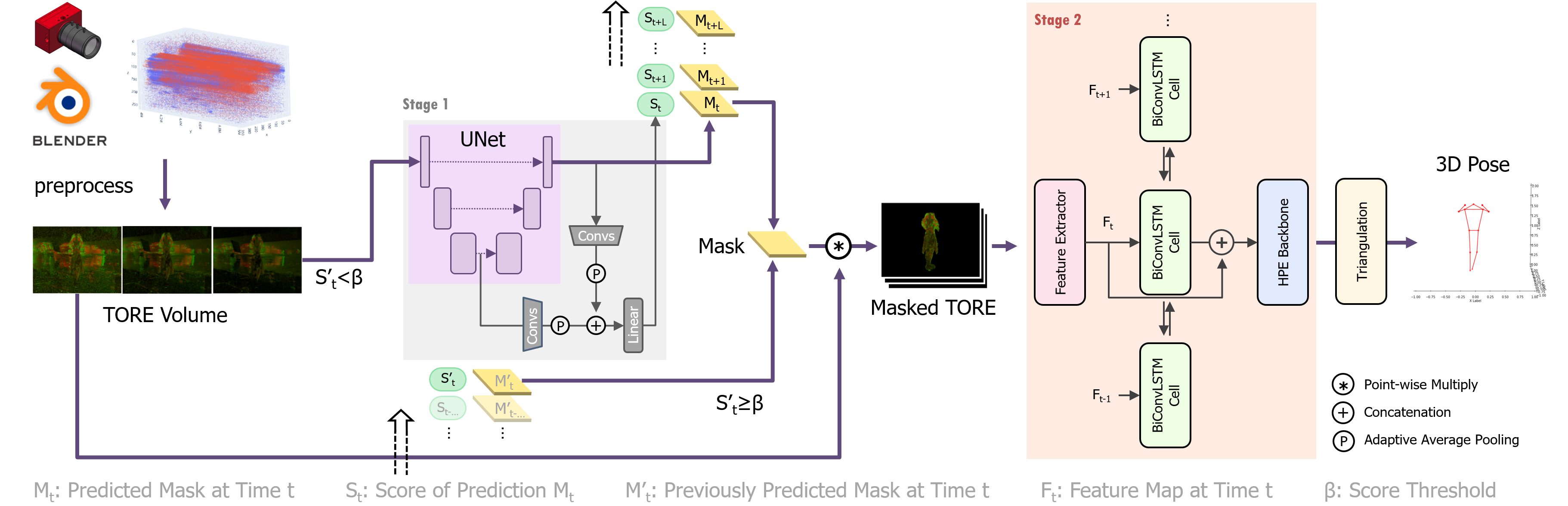}
    \caption{The pipeline of the proposed \emph{YeLan}. The event stream is processed into TORE volumes at first and then sent to the stage one human body mask prediction network. This network predicts a series of masks each time for the following frames with quality-assessment scores to reduce the computation cost. The estimated human mask is then point-wise multiplied by the original TORE volume before going to the next stage. Stage two is the human pose estimation network, where BiConvLSTM and three hourglass-like refine blocks are applied to estimate the heatmap of joints' projection on three orthogonal planes. The exact 3D coordinates of these joints are calculated by the triangulation method based on these heatmaps.}
    \label{fig:pipeline}
\end{figure*}

\subsection{Event Filtering with Human Body Mask Prediction Network}\label{unet}


In order to filter out events triggered by background activities and event camera hardware noise (e.g., hot pixel and leak noise \cite{Hu2021-v2e-cvpr-workshop-eventvision2021}), a mask prediction network is used that can predict a human body mask from the TORE volume representation.

For the mask prediction network, a modified version of U-Net \cite{ronneberger2015u} proposed by Olaf et al. is adopted. Unlike the original version, our modified version can predict the mask for the current input frame and a series of masks after this frame. Each predicted mask is generated together with a confidence score. The primary rationale behind predicting the human body mask of future frames is that future motion trajectories of different human body parts are generally predictable with information about current and previous motion trajectories. Time-Ordered Recent Event (TORE) volume representation efficiently captures current and previous motion history. It allows the mask prediction network to estimate human body masks of the current and future frames and their corresponding confidence scores. The confidence scores provide an early-exit-like mechanism where the computational pipeline bypass mask prediction of a certain frame if the mask predicted based on a previous frame's TORE volume achieves a high confidence score. This mechanism significantly saves computational cost and energy consumption (more results in section \ref{Evaluation}). For the predicted mask, the U-Net generates float numbers between 0 and 1, and a binarization process with a small threshold of $0.1$ is applied. The small threshold was used to ensure that the generated mask misses no parts of the human body since failing to cover the entirety will lead to more error than allowing a small amount of noise.

One major challenge for training the mask prediction network in an end-to-end scheme is the lack of a realistic event camera dataset with labeled human body mask sequences for recorded event streams. However, as our proposed motion-to-event simulator (described in section \ref{Simulator})  could generate paired pixel-level human masks with a high frame rate, they can be fed into the mask prediction network to train it fully.

\subsection{Human Pose Estimation Network}


The human pose estimation network consists of a ResNet-based feature extractor, a Bidirectional convolutional LSTM (BiConvLSTM) layer, an HPE backbone, and a triangulation module. The head part of ResNet34 serves as the feature extractor, followed by a BiConvLSTM layer with a skip connection. As the human body does not have abrupt changes during a relatively small time window, adjacent frames usually have similar ground truth labels. This continuity in human joint movements makes it helpful to refer to neighboring frames when estimating the joints' position. 
BiConvLSTM is a bidirectional version of ConvLSTM \cite{shi2015convolutional}, where ConvLSTM is a type of recurrent neural network for spatio-temporal prediction that has convolutional structures in both the input-to-state and state-to-state transitions. Lastly, the HPE backbone is made up of three hourglass-like CNN blocks. Each block outputs a series of marginal heatmaps to reconstruct the human joints' coordinates in 3D space. All the intermediate outputs from these three blocks are used to calculate the loss with the ground truth heatmaps, and the last two blocks could be considered refine networks. The feature extractor and the backbone network architecture have been developed based on a model proposed in \cite{scarpellini2021lifting}. For each joint, YeLan generates three heatmaps showing the probability of its projected position on \emph{xy}, \emph{xz}, and \emph{yz} planes. Then a soft-argmax operator is applied to extract the normalized coordinates of the joint. Lastly, predictions from the \emph{xy} plane are used as the final prediction for \emph{x} and \emph{y} coordinates, while values for \emph{z} are calculated by averaging the \emph{yz} and \emph{xz} predictions.

The ground truth labels used in the training and testing are normalized before being fed to the network. For a specific joint, we first project it to a plane parallel to the camera's image plane and have the same depth as the depth reference. The head joint's depth value is used as the depth reference. Then the 3D space in the DVS camera's view is mapped to a cube whose range is $[-1, 1]$. Lastly, as the network does not directly predict the 3D coordinates of a joint but predicts its marginal heatmaps instead, we also extract the joints' projection on three orthogonal faces of the normalized space cube to generate the ground truth for marginal heatmaps. The final marginal heatmaps are then calculated using a Gaussian filter on these projection images \cite{scarpellini2021lifting}.

\subsection{Loss}
For the mask prediction network, the loss function consists of three components. The first component is the Binary Cross Entropy(BCE) loss calculated between the predicted mask series $\hat{M}$ and the corresponding ground truth masks $M$. This loss is applied to guarantee the accuracy of all the generated masks. Next, a Mean Square Error(MSE) loss is calculated over the predicted confidence scores $S$ and their ground truth. The ground truth score is the Mean Absolute Error(MAE) between a predicted mask and its ground truth mask. Lastly, although we want to predict the mask series for the current frame and frames following it, they are not equally important. We need to guarantee that closer frames' masks get weighted more, especially for the current frame. Therefore, another BCE loss is calculated between the predicted mask for the current frame ($\hat{M}_0$) and its corresponding ground truth $M_0$. The overall loss is:
\begin{equation}
\begin{split}
    Loss_{mask} =& BCE(M, \hat{M}) + BCE(M_0, \hat{M}_0) +\\
                 &MSE(S, MAE(M, \hat{M}))
\end{split}
\end{equation}

We adopted the loss introduced in \cite{scarpellini2021lifting} for the human pose estimation network. As the marginal heatmaps can be interpreted as probability distributions of joint locations, Jensen-Shannon Divergence(JSD) can be applied between the predicted heatmaps by each block ($\hat{H}^i$, where $i$ represents the block index) and the ground truth heatmaps $H$ on each projection plane ($xy, xz, zy$). Also, a geometrical loss is calculated between the reconstructed 3D joints' coordinates $\hat{p}_{xyz}$ and their ground truth $p_{xyz}$. The loss for this stage can be written as follows:
\begin{equation}
\begin{split}
    Loss_{HPE} &= \sum_i(||\hat{p}^i_{xyz}-p_{xyz}||_2+JSD(H_{xy}, \hat{H}_{xy})+\\
               &JSD(H_{xz}, \hat{H}_{xz})+JSD(H_{zy}, \hat{H}_{zy}))
\end{split}
\end{equation}

\section{Synthetic Data Generation with Comprehensive Motion-to-Event Simulator}\label{Simulator}

First of all, although there are some event camera-based HPE datasets \cite{Calabrese2019DHP19DV}, these datasets mainly focus on fixed everyday movements (e.g., walking, jumping, waving hands), making models trained on that fail when encountering complex movements. A dance performance will generally include fast and complex gestures, which are rare in a general everyday gesture dataset such as \cite{Calabrese2019DHP19DV}. Moreover, these datasets have been collected in ideal lighting conditions with blank or static backgrounds that do not represent the real-world environment. Moreover, the existing dataset suffers from the lack of diversity in participants, motion dynamics, clothing style, and types of background activities. For clothes, all the participants wear the same black close-fit clothes. For motion, the participants follow a fixed set of pre-defined activities, which are relatively standardized, simple, and repetitive. However, in dance, human moves in way more dynamic, complex, and challenging patterns. To bridge this critical gap, we propose to generate synthetic data with a comprehensive motion-to-event simulator.

As for simulators, although there are some existing event camera simulators  \cite{mueggler2017event, rebecq2018esim, Hu2021-v2e-cvpr-workshop-eventvision2021, joubert2021event}, there is a common and crucial problem. Almost all of them only work to turn existing pictures or videos into event streams instead of making highly customized event streams directly based on the research problem needs. Moreover, for existing simulators, if the original video does not come with human joint labels, the generated event stream is also label-less. However, in this work, the simulator we build adapts a physics-aware rendering system and makes all the concerned parameters fully controlled and customized, including lighting, motion, human gender, body shape, skin color, clothes and accessories, background, and scenes. This enables us to apply complex dance movements, and all the data we generate is paired with accurate labels. 



\begin{figure*}[t]
    \centering
    \includegraphics[width=\linewidth]{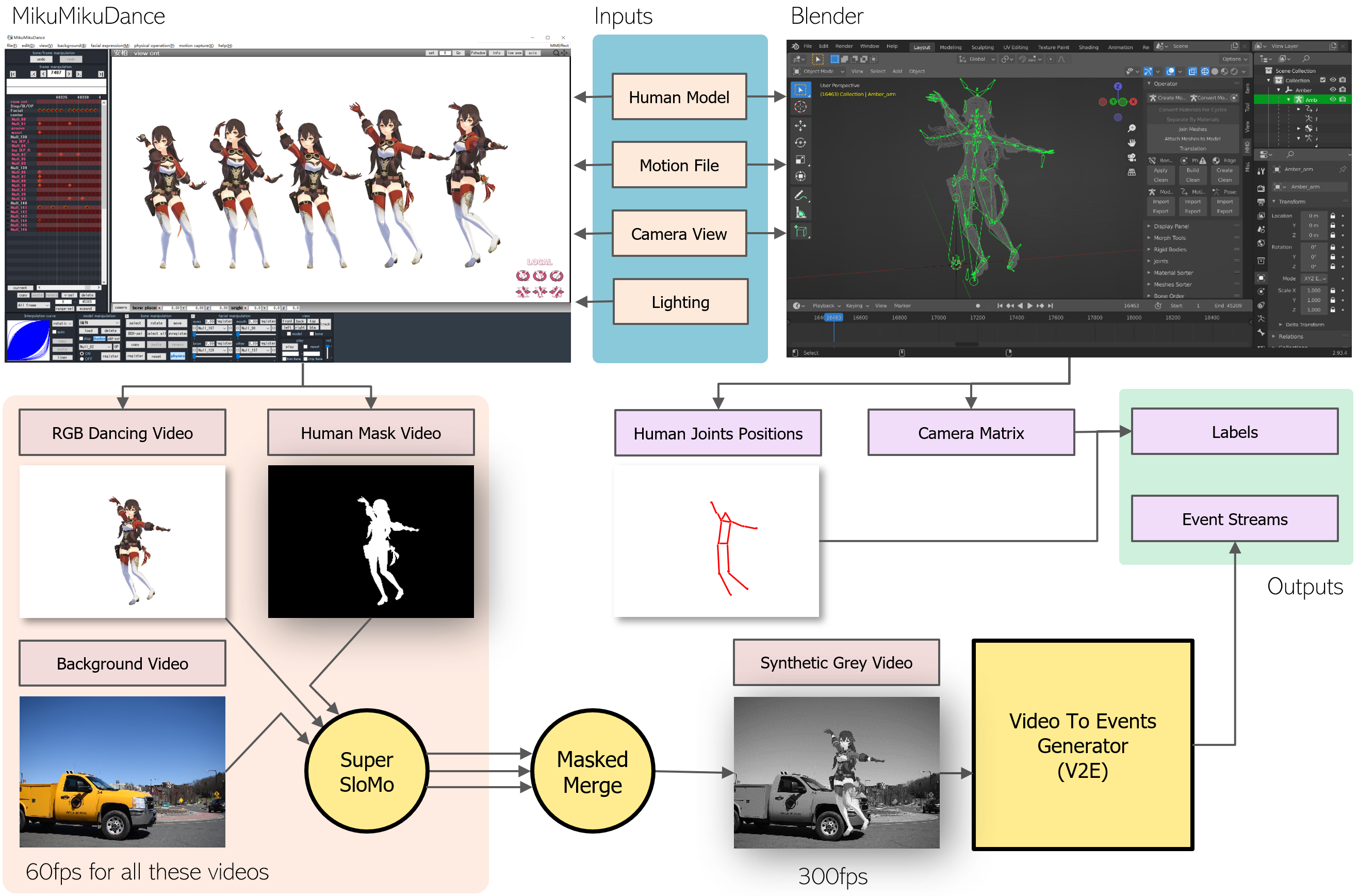}
    \caption{The pipeline of synthetic data generation with the comprehensive motion-to-event simulator. Motion file, human model, camera view, lighting, and other settings are the input to our motion-to-event simulator pipeline. These inputs are rendered into RGB and human mask videos in MMD and merged with background videos. Merged videos are processed as event streams using V2E, and the corresponding labels are calculated in Blender.}
    \label{fig:syn_pipeline}
\end{figure*}

\subsection{Advantages of Synthetic Data}

Several significant advantages of using a synthetic data generator are highlighted by a recent work \cite{Goyal2022VisionMA}. Real-world data collection is expensive and time-consuming. The cost per participant is high, limiting the sample size, the scale of scenes, and the diversity in physical and environmental characteristics. The synthetic dataset generation process is fully controllable. The human models, background scenes, movements, lighting conditions, blur scale, camera distance, and camera view angle can all be precisely controlled, making it practical to modify a specific condition while keeping all the other factors unchanged. The output video file can be rendered with a very high frame per second (FPS) without blur problems caused by under or over-exposure in dynamic lighting conditions. Lastly, the ground truth of human joints' 3D location can be extracted inside the software for different dances and human models.



\subsection{Tools used in the simulator}

Synthetic data generation consists of RGB dance video rendering, human joints' position extraction, and events generation. MikuMikuDance (MMD), Blender, and V2E are chosen for each step, respectively.

\emph{MMD} is a freeware animation program that lets users animate and creates 3D animated movies. This software is simple but powerful, with a long history and a big open-source community behind it. Plenty of human models, scenes, and movement data can be easily accessed for free. In addition, it can automatically handle clothing physics and interaction with the body in a decent manner with minimal manual adjustment.

Software \emph{Blender} is adopted to generate human joints' ground truth labels and camera matrix. \emph{Blender} is a free and open-source 3D computer graphics software for creating animated films, motion graphics, etc. It is highly customizable, and all essential information can be accessed during rendering, including the 6-degrees of freedom coordinates of human joints and the camera center. The 13 key points' ground truth coordinates are extracted at 300 frames per second (FPS).

Lastly, \emph{Video to Event (V2E)} is used for event generation. V2E is a toolset released in 2021 by Delbruck et al. \cite{Hu2021-v2e-cvpr-workshop-eventvision2021}. It can synthesize realistic event data from any conventional frame-based video using an accurate pixel model that mimics the event camera's nonidealities. According to its author, V2E supports an extensive range of customizable parameters and is currently the only tool to model event cameras realistically under low illumination conditions.


\begin{figure*}[t]
    \centering
    \includegraphics[width=\linewidth]{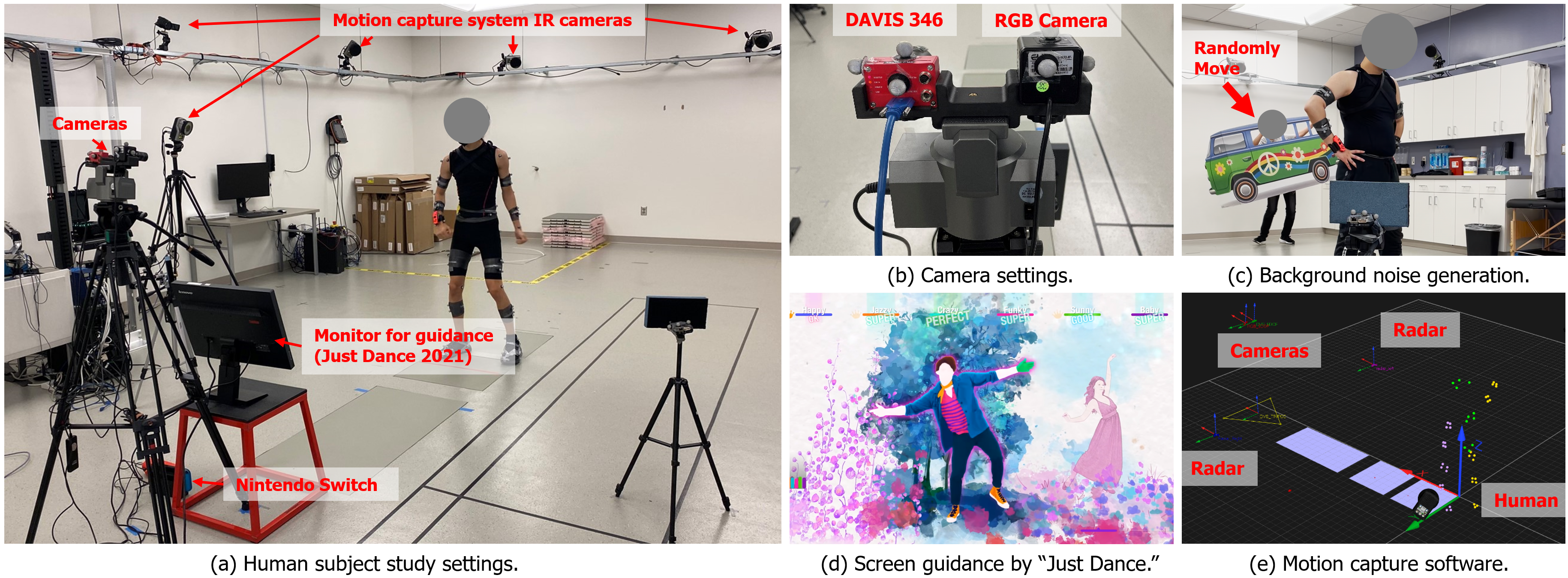}
    \caption{Settings of human subject data collection in an indoor laboratory setting.}
    \label{fig:NiDHP_overview}
\end{figure*}

\subsection{Comprehensive Motion-to-Event Simulator}

As Fig. \ref{fig:syn_pipeline} shows, our proposed motion-to-event generator takes 3D character models, motion files, camera views, and lighting conditions as inputs. MMD renders RGB dance videos and their paired human mask videos given these inputs, while Blender generates corresponding ground truths. For each camera view, a camera intrinsic and extrinsic matrix is calculated by Blender as well.

Then our simulator combines these rendered dance videos with collected background videos by referring to the paired mask videos. According to \cite{Hu2021-v2e-cvpr-workshop-eventvision2021}, if a video's temporal resolution is low, generated event stream will be less realistic. However, due to the software limitation and background video quality, the synthesized videos are at 60 FPS. This gap in FPS is compensated using SuperSlowMo\cite{jiang2018super}, which can interpolate videos to a high FPS with convincing results. To reduce the time and computation cost, we only apply SuperSlowMo to dance and background videos before the merging. This way, we do not have to interpolate synthesized videos for each human and background combination.

After the RGB video rendering, merged videos are sent to the V2E events stream generator. Many parameters, like event trigger thresholds, noise level, and slow-motion interpolation scale, can be modified in detail. These features enable us to generate many highly customizable DVS event streams at a meager cost in a short time. To simulate situations in the real world, we increase the noise as the brightness decreases. 



For human joints' ground truth, as mentioned above, we write our customized scripts and inject them into Blender to collect the exact position of all joints while rendering scenes at 300 FPS. Also, scripts help extract the camera's intrinsic and extrinsic matrix used in the label pre-processing. Besides the advantages mentioned above, 3D human models have even more flexibility initially. Skin color, height, body style, clothing, hair color and style, and accessories are all easily modifiable - which is very difficult to do in real-world data collection. 

\begin{figure*}
    \centering
    \includegraphics[width=\linewidth]{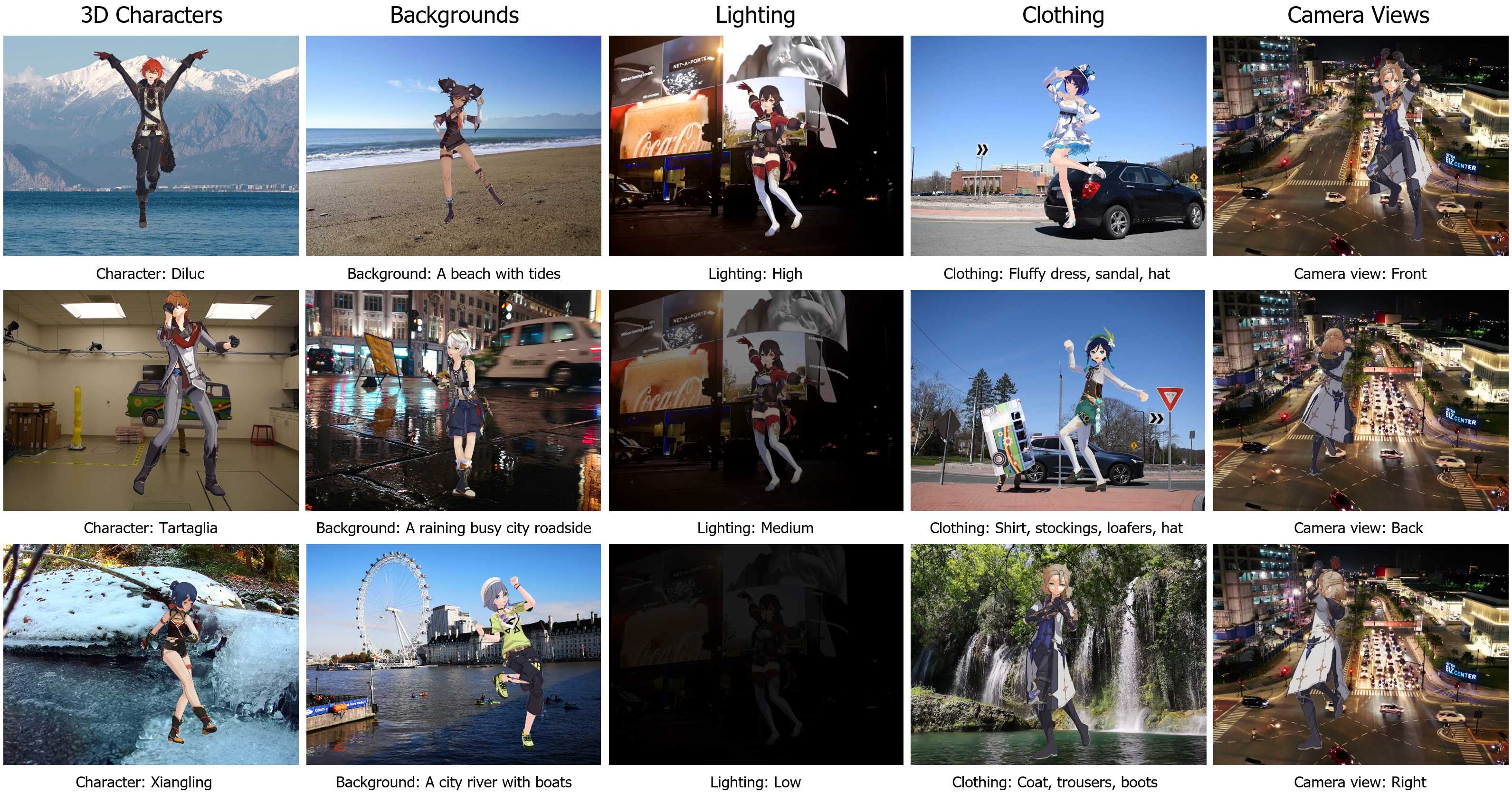}
    \caption{RGB frame samples from the synthetic data generated by the comprehensive motion-to-event simulator.}
    \label{fig:MiDHP_samples}
\end{figure*}

\subsection{Synthetic Dataset Description}
With the comprehensive motion-to-event simulator, we have generated about 4 million data pieces (more specifically, 3,958,169 snippets). Examples of the synthetically generated RGB frames are shown in figure \ref{fig:MiDHP_samples}. The total dataset size is about three terabytes. This data was synthesized from  1320 combinations of a few different variables, including 10 human models, 8 pieces of one-minute dance motions, 11 background videos, 4 camera views (i.e., front, back, left, right), and 3 different lighting conditions (i.e., high, medium, low). This dataset contains processed TORE volume, paired labels and masks, constant-time frames with the same time step (20ms), and raw event stream files for generating any other customized representations. Due to the computational resources and training time limitation, only 30\% of data instances from selected 330 setting combinations are used for training and validation. They are shuffled and divided with a ratio of 8:2. Then we randomly select 82 setting combinations from the rest 990 unused combinations as the test set.

\section{Human Subject Study with Dynamic Lighting and Background}

\subsection{Real-World Dataset in Motion Capture Facility}
In addition to the synthetic dataset, we have conducted an Institutional Review Board (IRB)-approved human subject study to collect a real-world human pose dataset from nine participants. Seven of them are male, and the other two are female. The participants were 20-31 years old and recruited from a university campus using a snowball sampling technique. During this study, our participants were asked to play a Nintendo dance game ``Just Dance 2021''. Each participant selected five songs to dance to during the study after a short training period with the tutorial dances in the Nintendo dance game. A monitor was used to provide the participants with further guidance/cues for participants to follow along.


A 9-camera-based 3D motion capture system (Qualisys AB, Göteburg, Sweden) is used to obtain ground truth 3D kinematics data of the human body, which provides us with the 3D absolute coordinates for all selected human joints at a frame rate of 200 frames per second (FPS). An event camera (DAVIS 346 \cite{Brandli2014A2}) and an RGB camera are simultaneously run to record the participants' movements. Other necessary equipment includes a flicker-free LED light, IR filter, cardboard background, Monitor, and Nintendo Switch. During the dataset collection, the lighting condition was strictly controlled to induce low-lighting and high-lighting conditions. All the lights except a dim flicker-free lamp are turned off during the low-lighting conditions session. The IR filter is attached in front of the event camera's lens to filter out the events caused by IR light emitted by the motion capture system. Figure \ref{fig:NiDHP_overview}  illustrates the data collection settings, hardware, sensor arrangements, and mechanisms to generate background noise.

\begin{figure*}
    \centering
    \includegraphics[width=\linewidth]{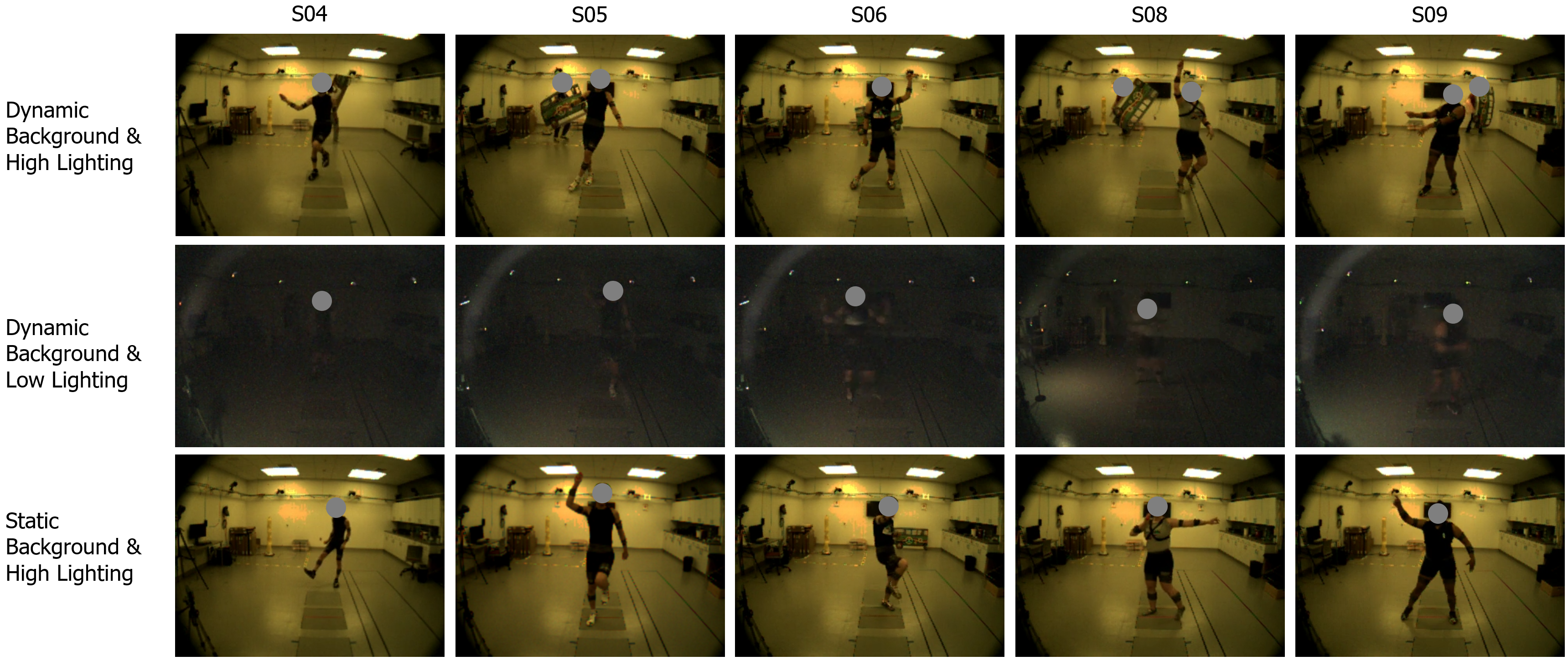}
    \caption{RGB frame samples from the real-world dataset collected in motion capture laboratory.}
    \label{fig:NiDHP_samples}
\end{figure*}

In the dynamic background condition, a person behind the target participant moved randomly with a large cardboard (that depicts a bus or car). For static background cases, on the other hand, no other movable contents appear in the background during recording. On average, each participant danced for about 20 minutes, and the time was distributed equally to the four conditions sessions mentioned above. We divided the real-world dataset simply by selecting all the events generated by participants seven and eight as test sets. Participant seven is male, and eight is female. All other data is shuffled and divided into training and validation sets with the same 8:2 ratio. We prepared a short video To illustrate further the rich synthetic and real-world data, which can be found at \url{bit.ly/yelan-research}.

\section{Experiments and Results}\label{results}

\begin{table*}[ht]
\caption{Test set results comparison on three metrics for synthetic and real-world datasets.}
\begin{tabu}{l l l l l l l l}
\tabucline[2pt]{-}
& & \multicolumn{3}{c}{\textbf{Synthetic}} &  \multicolumn{3}{c}{\textbf{Real-world}}\\ \hhline{~~------}
Methods                                     & Representation       & MPJPE$\downarrow$ & PCK(\%)$\uparrow$ & AUC(\%)$\uparrow$ & MPJPE$\downarrow$ & PCK(\%)$\uparrow$ & AUC(\%)$\uparrow$ \\ \hline
Scarpellini et al., 2021                         & Constant Time & 91.88 & 83.92 & 80.95 & 111.57 & 78.12 & 77.03                    \\ \hline
Baldwin et al., 2021                & TORE        & 59.34 & 93.17 & 86.97 & 101.967 & 82.22 & 78.90                    \\ \hline
Our Model   & TORE       & \textbf{46.57} & \textbf{96.34} & \textbf{89.37} & 96.61 & 81.88 & 79.78                   \\ \hline
Our Model w/i pre-training  & TORE       & - & - & - & \textbf{90.94} & \textbf{85.14} & \textbf{80.91}                   \\
on synthetic \& fine-tuning &&&&&&& \\
on real data   &       &  &  &  &  &  &  \\
\tabucline[2pt]{-}
\end{tabu}
\label{tab:baseline}
\end{table*}


\subsection{Training Details}

All models are trained on eight 1080ti or 2080ti graphic cards. The batch size is identical over all the training: Stage one has a batch size of 128, and stage two has a batch size of 16. All stages are trained with the following parameters: 0.001 learning rate, Adam optimizer with $1e{-5}$ weight decay, early stopping with the patience of 10 epochs. A learning rate scheduler is applied to halve the learning rate every $N$ epoch. $N$ is set to 5 in stage one and 10 in stage two, considering the training pattern and convergence speed difference. This project is implemented in Pytorch-Lightning. Stage one and two in \emph{YeLan} are trained separately. For the mask prediction network, due to the ground truth limitation, it is only trained on synthetic data and used directly in both synthetic and real-world datasets. For stage two, it is pretrained on synthetic dataset at first, and then fine-tuned on real-world dataset.


\subsection{Evaluation metrics}
\begin{figure*}[ht]
    \centering
    \includegraphics[width=\linewidth]{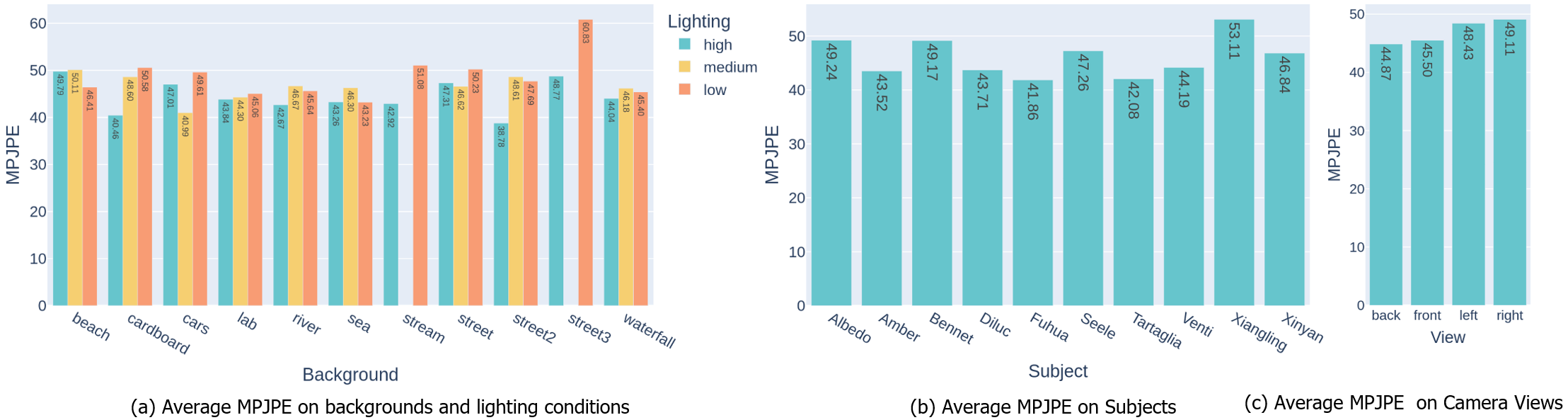}
    \caption{The performance across lighting conditions, dynamic backgrounds, subjects, and camera views in our synthetic data.}
    \label{fig:statistics_syn}
\end{figure*}

Following the convention, we used Mean Per-Joint Position Error(MPJPE), the most commonly used metric in HPE, as our primary evaluation metric. Another two popular metrics, PCK and AUC, are also compared in the major comparison. MPJPE is calculated by averaging the Euclidean distance between each predicted joint and its corresponding ground-truth joint, usually at the millimeter level. PCK stands for Percentage of Correct Keypoints according to a threshold value of 150mm, which is commonly used. If the predicted joint is within a 150mm cube centered at the ground truth joint, it is treated as correctly predicted and returns 1; otherwise, it returns 0 for the wrong case. AUC means the Area Under the Curve for the PCK metric at different thresholds. We also use the standard threshold sets, 30 evenly spaced numbers from 0 to 500mm. When calculating the AUC, we will first calculate the PCK at all these threshold values, and their mean value is the target AUC.

\begin{align}
    MPJPE &= \frac{1}{J} \sum^J_i||p_{xyz}-\hat{p}_{xyz}||\\
    PCK_\alpha &= \frac{1}{J} \sum^J_i sign(\alpha-||p_{xyz}-\hat{p}_{xyz}||)\\
    AUC &= \frac{1}{N} \sum_{\alpha=0}^A PCK_\alpha(p_{xyz}, \hat{p}_{xyz})
\end{align}

Where $p_{xyz}$ and $\hat{p}_{xyz}$ represent the ground truth and predicted 3D joint position, and $J$ means the number of skeleton joints. In PCK and AUC's formula, $\alpha$ and $A$ mean the threshold and the maximum threshold in AUC calculation. $N$ in AUC stands for the number of the different thresholds used.

\subsection{Baselines}

\begin{figure*}
    \centering
    \includegraphics[width=\linewidth]{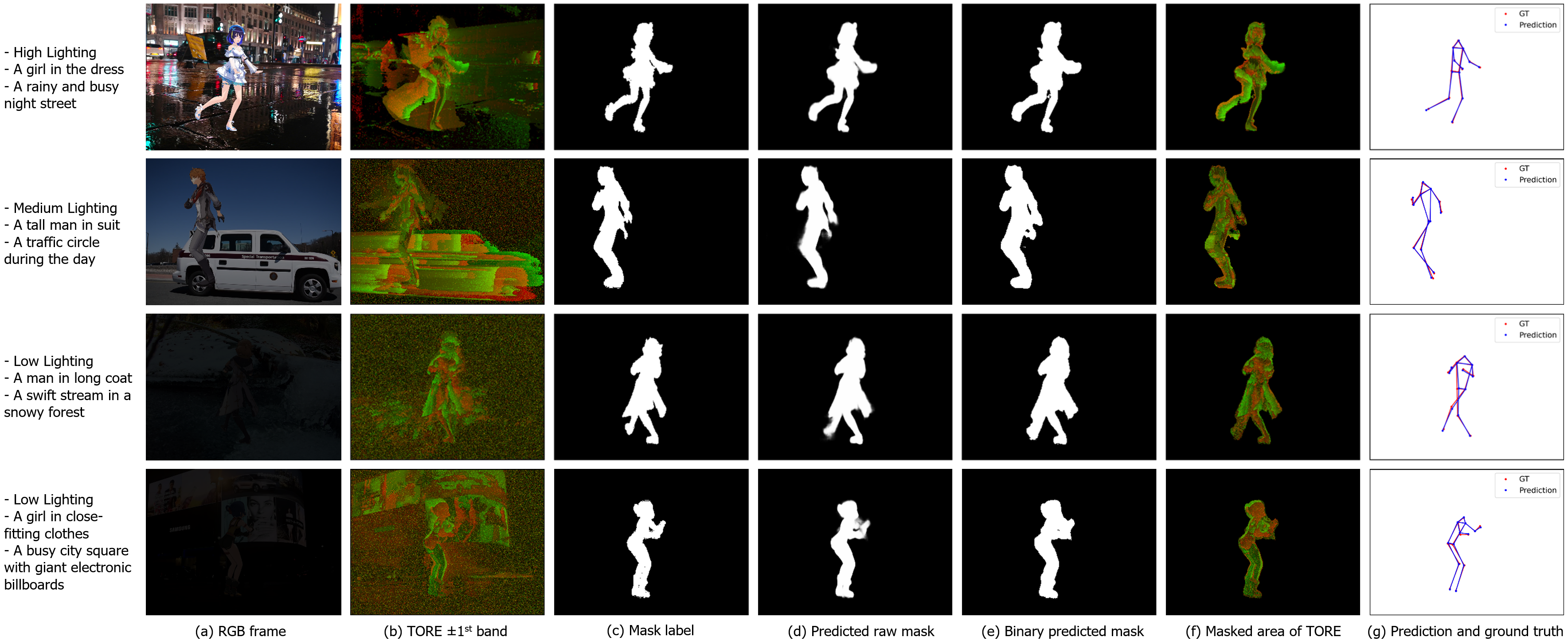}
    \caption{Sample results from our synthetic dataset in different lighting conditions with dynamic backgrounds. The RGB frames, corresponding event representation TORE, generated masks, ground truth, and predicted 3D human pose are shown.}
    \label{fig:results_with_mask}
\end{figure*}

The baseline methods selected for this work are \cite{scarpellini2021lifting} and \cite{Baldwin2021TimeOrderedRE}, proposed by Scarpellini et al. and Baldwin et al. (originally published in 2021). For \cite{scarpellini2021lifting}, They used two event-count-based representations: constant event count frames and voxel grid, which have a variable time step and thus lose synchronization with our labels and TORE volumes.  Therefore, the original work is modified to take constant time representations as input instead. The time step for this representation is set to 20 ms as well to guarantee that the labels are matched. Baldwin et al. proposed TORE volume, and they evaluated on the model and dataset proposed in \cite{Calabrese2019DHP19DV} with a replaced representation to show their new representation's superiority on multi-view human pose estimation task. Since in this paper we are interested in developing a monocular human pose estimation framework, we use the same monocular HPE model in \cite{scarpellini2021lifting} with the representation switched to TORE to show the performance of \cite{Baldwin2021TimeOrderedRE} in this new task.

\subsection{Evaluation}\label{Evaluation}

Table \ref{tab:baseline} illustrates the performance comparison between the proposed \emph{YeLan} and baseline models on both synthetic and real-world datasets. The table shows that the baseline models consistently underperform the proposed YeLan model in both conditions. In the synthetic dataset, the model proposed by the Scarpellini et al. \cite{scarpellini2021lifting} achieves an MPJPE and PCK of respectively 91.88 and 83.92\% while the model proposed by the Baldwin et al. \cite{Baldwin2021TimeOrderedRE} with TORE representations have a slightly improved performance of respectively 59.34 MPJPE and 93.17\% PCK. The proposed \emph{YeLan} system outperforms the baseline models by a significant margin and achieves the lowest MPJPE of 46.57 and the highest PCK of 96.34\% on the synthetic dataset. 

Compared to the synthetic data, the performance of all the models on the real-world data is significantly worse. For example, while our proposed model (trained and tested on the same type of data) achieves an MPJPE of 46.57 with the synthetic data, the performance in the real-world dataset is 96.61 MPJPE. The real-world data comes with its own challenges and unique characteristics. Real human motion trajectories are slightly different from simulated trajectories. There is also substantial heterogeneity in the skeletal formations and impedance matching states in the real-world data. Lastly, another researcher is inside the view generating background activities, who sometimes get recognized and masked as an additional human, which could confuse the models.
Consequently, the performance of the real-world data will be expected to be lower than that of the synthetic data. However, the proposed \emph{YeLan} still achieves the lowest metrics compared to the baseline models. The baseline models are designed to work in ideal environments which do not suffer from dynamic lighting conditions and moving background content problems, which contributes to their poor performance in more realistic environments.

As Fig. \ref{fig:statistics_syn} shows, \emph{YeLan} has strong stability over the changes in lighting conditions and camera views, while more complicated background contents cast an impact on the results. Different human models also show an impact on performance. From the statistics on human models, we know that human models ``Albedo", ``Seele," and ``Xiangling" performs less well than other human models. By looking at the original 3D model and test data composition, the reason becomes obvious: Albedo wears a long coat reaching his knee, while Seele wears a fluffy dress with a complex structure. These factors make their joint position harder to estimate. As for Xiangling, though she does not have clothing-related problems, we find some body deformation that happened during the simulation. Also, the randomly selected test cases for Xiangling contain more difficult backgrounds and lighting combinations, such as two busy city street environments in low-lighting conditions, according to Fig. \ref{fig:statistics_syn} (a).

\begin{figure}[h]
    \centering
    \includegraphics[width=\linewidth]{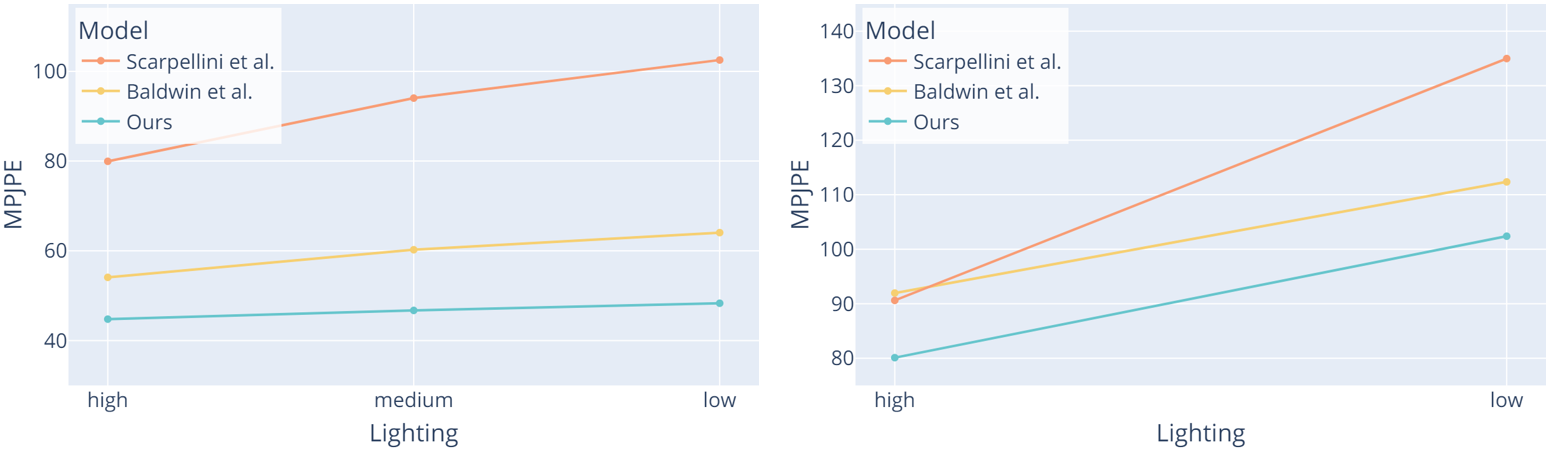}
    \caption{Impact of lighting condition on the performance (MPJPE) on our synthetic (left) and real-world (right) dataset.}
    \label{fig:comp_light}
\end{figure}

Moreover, Fig. \ref{fig:comp_light} shows that our model behaves better constantly in all lighting conditions compared to baseline models. This feature is significant in real-world applications as the pipeline could generate convincing predictions regardless of the lighting conditions.

\begin{figure}[ht]
    \centering
    \includegraphics[width=0.6\linewidth]{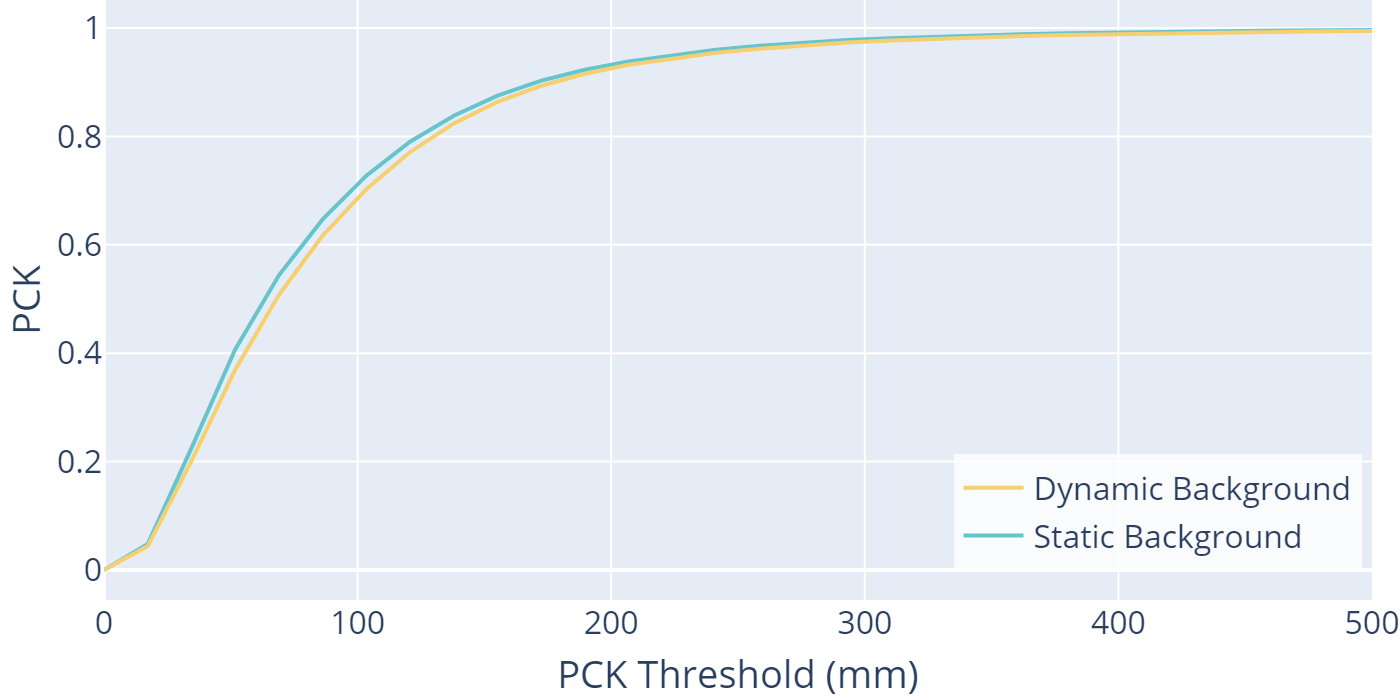}
    \caption{AUC curve with dynamic and static background on our real-world dataset.}
    \label{fig:static_dynamic_bkg}
\end{figure}

The mask prediction network in \emph{YeLan} enables it to reach a similar level of PCK and AUC when the background is dynamic and static. We compared on NiDHP test set with dynamic and static background cases using the model pretrained on MiDHP and fine-tuned on NiDHP. Fig. \ref{fig:static_dynamic_bkg} depicts that \emph{YeLan} get a comparable result on dynamic background cases with static ones.

\subsection{Ablation Study}

As introduced in section \ref{pipeline}, there are two most essential modules in \emph{YeLan}: a mask prediction network and the BiConvLSTM. We do an ablation study by removing one module each time and comparing their performance on the synthetic dataset. If the mask prediction network is removed, the resulting MPJPE is 63.99; if the BiConvLSTM is removed, the corresponding MPJPE is 49.08. In contrast, the complete YeLan pipeline gets an MPJPE of 46.57, which proves the effectiveness of these modules.




\subsection{Impact of Pretraining on Synthetic Data}




Synthetic data from the physics-based simulator is a core contribution of this work which allows us to generate event streams from diverse virtual 3D characters in different simulated settings. They include different clothing, lighting conditions, camera viewing angle, dynamic background, and movement sequence. The generated synthetic data is faithful to physics and can provide essential data for pretraining \emph{YeLan} since running real-world experiments with too many participants is expensive and time-consuming. To this end, we hypothesize that the pretraining on synthetic data will allow \emph{YeLan} to learn better feature representation of physical conditions in diverse simulated settings and achieve superior performance on the real-world dataset.

Table \ref{tab:baseline} demonstrates that our proposed model achieves superior performance (concerning the MPJPE/PCK/AUC metrics) on the real-world dataset if the model is pretrained on the synthetic dataset from the simulator. The model trained only on the real-world dataset achieves 96.61 MPJPE and 81.88\% PCK. On the other hand, if the model is pretrained on the synthetic data and then fine-tuned on the real-world dataset, it achieves 90.94 MPJPE and 85.14\% PCK, which clearly shows the benefit of pretraining on synthetic data from a simulator.


\subsection{Occlusion}

\begin{figure}[h]
    \centering
    \includegraphics[width=0.8\linewidth]{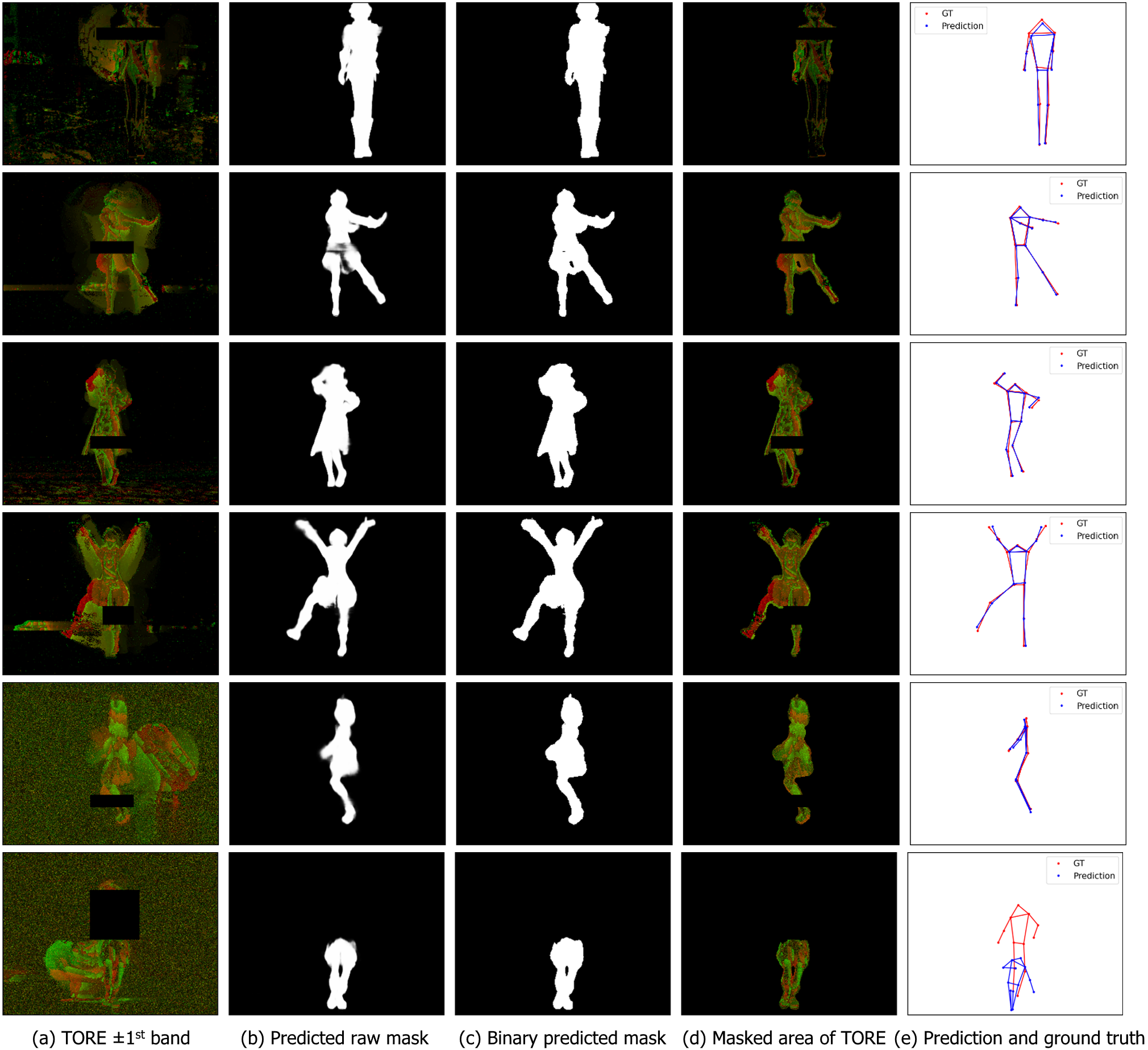}
    \caption{Samples of occluded inputs, predicted mask, TORE volume representation and inferred human poses.}
    \label{fig:occlusion}
\end{figure}

In the real world, occlusion is also an inevitable problem. In our synthetic dataset, different types of clothes and accessories like hats, long hair, fluffy skirts, and long coats already occluded the human body to a relatively large extent. Moreover, the camera's side views also introduced many self-occlusion. The excellent performance over all these scenarios shows a solid ability to survive occlusion.

To further prove the occlusion resiliency of the system, we augmented the dataset with more block occlusions. In the eye of the event camera, if a static object is placed in front of a human and shadows him, the corresponding area shoots no event due to the object's immobility. Regarding the input TORE volume, the occluded area becomes pure black, as no event activity is recorded. To simulate the occlusion like this, we trained a model with the value of random areas set to zero. These rectangular occlusion areas have random sizes and locations, and the occlusion is also applied randomly with a probability of 80\% during the training. When testing on the test set with random occlusion enabled, the MPJPE is 96.794. During the test, the occlusion probability is set to 100\%. There is an accuracy drop, but considering that the maximum random occlusion area is $80\times80$ (where the frame size is $260\times346$), it makes sense as sometimes the majority of the human body could be occluded. Fig. \ref{fig:occlusion} shows samples from the test set.

\subsection{Systems Benchmarking}

\begin{figure}[ht]
    \centering
    \includegraphics[width=0.6\linewidth]{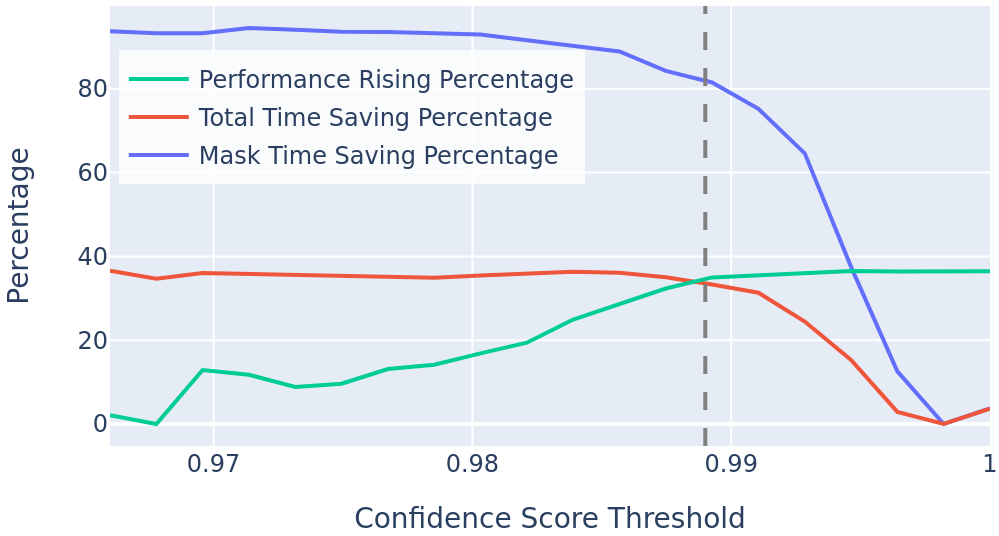}
    \caption{Time-saving and performance rising percentage with regard to confidence score threshold.}
    \label{fig:score-threshold}
\end{figure}

As is introduced in \ref{unet}, the first stage of \emph{YeLan} is an early-exit-style human mask prediction network, where a threshold $\beta$ is used to decide when to start a new inference. The selection of $\beta$ is a trade-off. If the threshold is set too high, the mask prediction process will be executed too many times, resulting in a longer inference time. On the contrary, many defective masks will be used if the threshold is too low, which harms the overall accuracy. In order to select the best threshold, we ran an experiment to do human pose estimation on a continuous one-minute event stream with a series of different thresholds. The accuracy and time consumption are recorded and made into Fig. \ref{fig:score-threshold}. From this figure, we can observe that as we increase the threshold, accuracy goes up and time consumption goes down, while we can achieve a balanced accuracy and time consumption at near 0.99.

\begin{figure}[ht]
    \centering
    \includegraphics[width=0.6\linewidth]{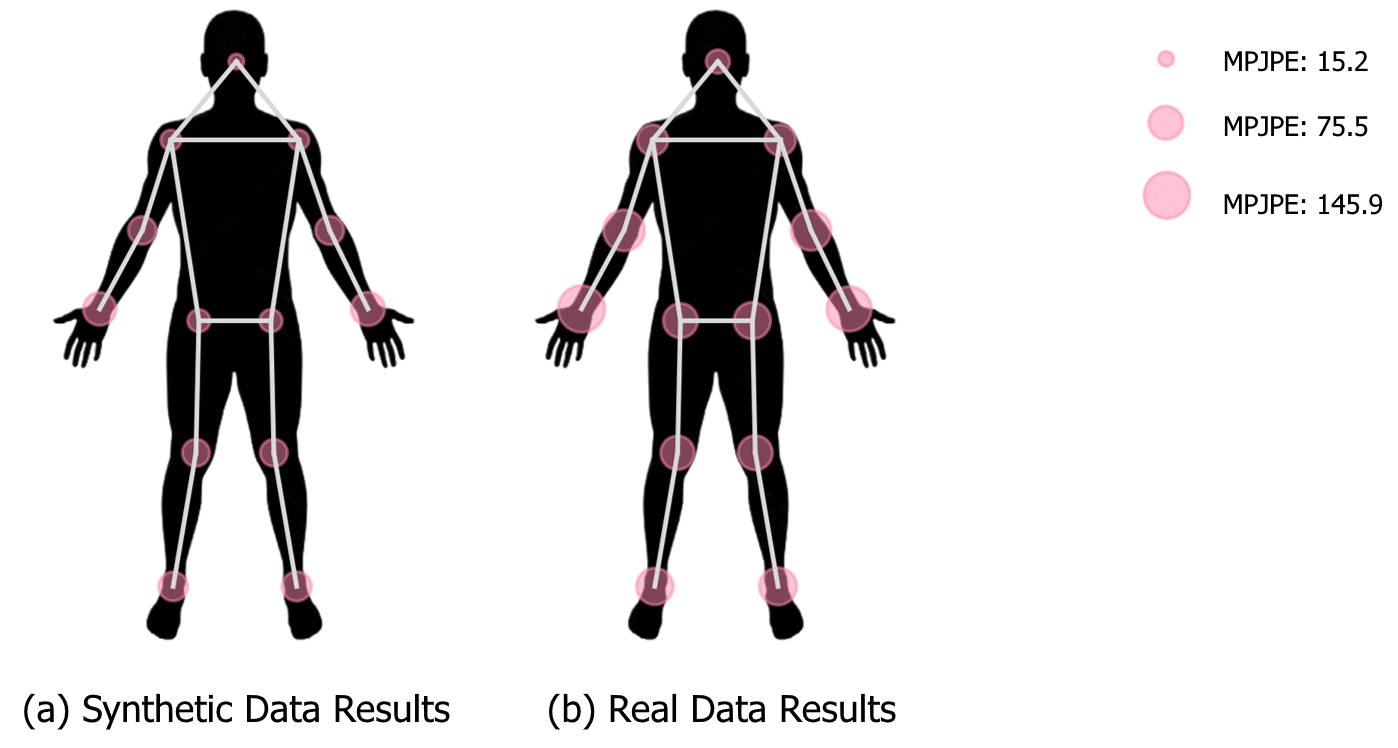}
    \caption{Joint-wise average MPJPE analysis on our synthetic dataset.}
    \label{fig:joint-mpjpe}
\end{figure}

Moreover, to understand the error distribution across different joints, we calculated the mean MPJPE for all 13 joints. As Fig. \ref{fig:joint-mpjpe} shows, the head joint has the lowest MPJPE, and the shoulder joints enjoy the second-lowest error. On the other hand, the left and right hands have the highest MPJPE, and foot joints come next. From our observation, this happens because the head and shoulders share the most stable contours across different characters and overlap less with other body parts. However, as hands and feet locates farthest from the human center and move the most during the dancing, they are harder to estimate due to more possible movements and patterns. More patterns lead to more uncertainty, which harms the accuracy even more when occlusion happens.

The total model size of \emph{YeLan} is 234 MB. As for the time consumption, we run a full test on the MiDHP test set with a batch size of one, and the average time cost on each frame is about 29.1 ms when running on a 2080ti graphic card.

\section{Comparison between Different Modalities}
\subsection{RGB Camera}

\begin{figure*}[h]
    \centering
    \includegraphics[width=\linewidth]{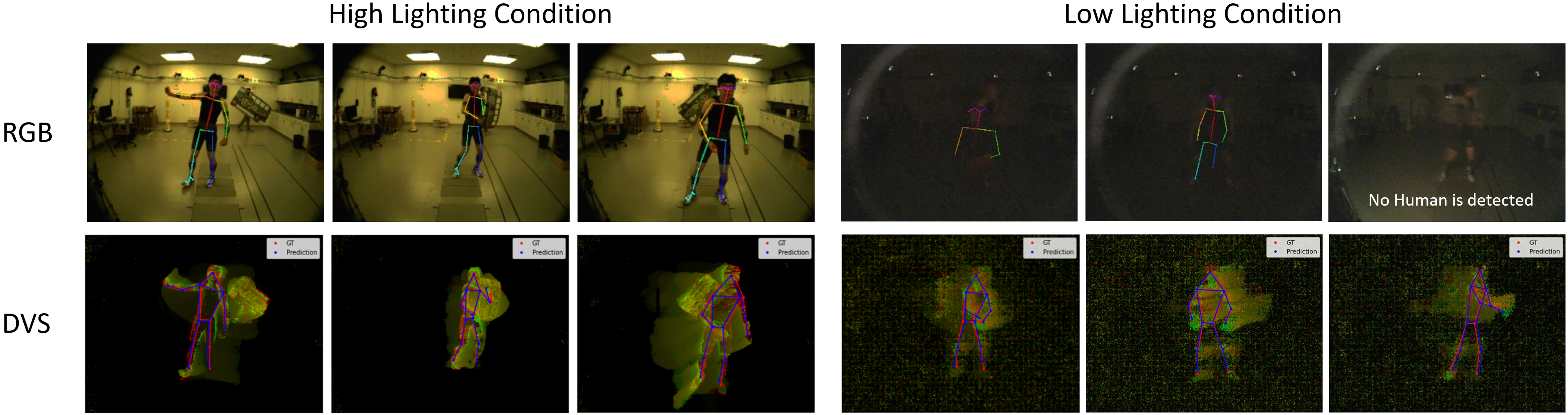}
    \caption{HPE comparison between RGB and DVS in both high and low-lighting conditions. RGB and DVS are synchronized.}
    \label{fig:rgb_dvs_comp}
\end{figure*}

\begin{figure}[h]
    \centering
    \includegraphics[width=0.6\linewidth]{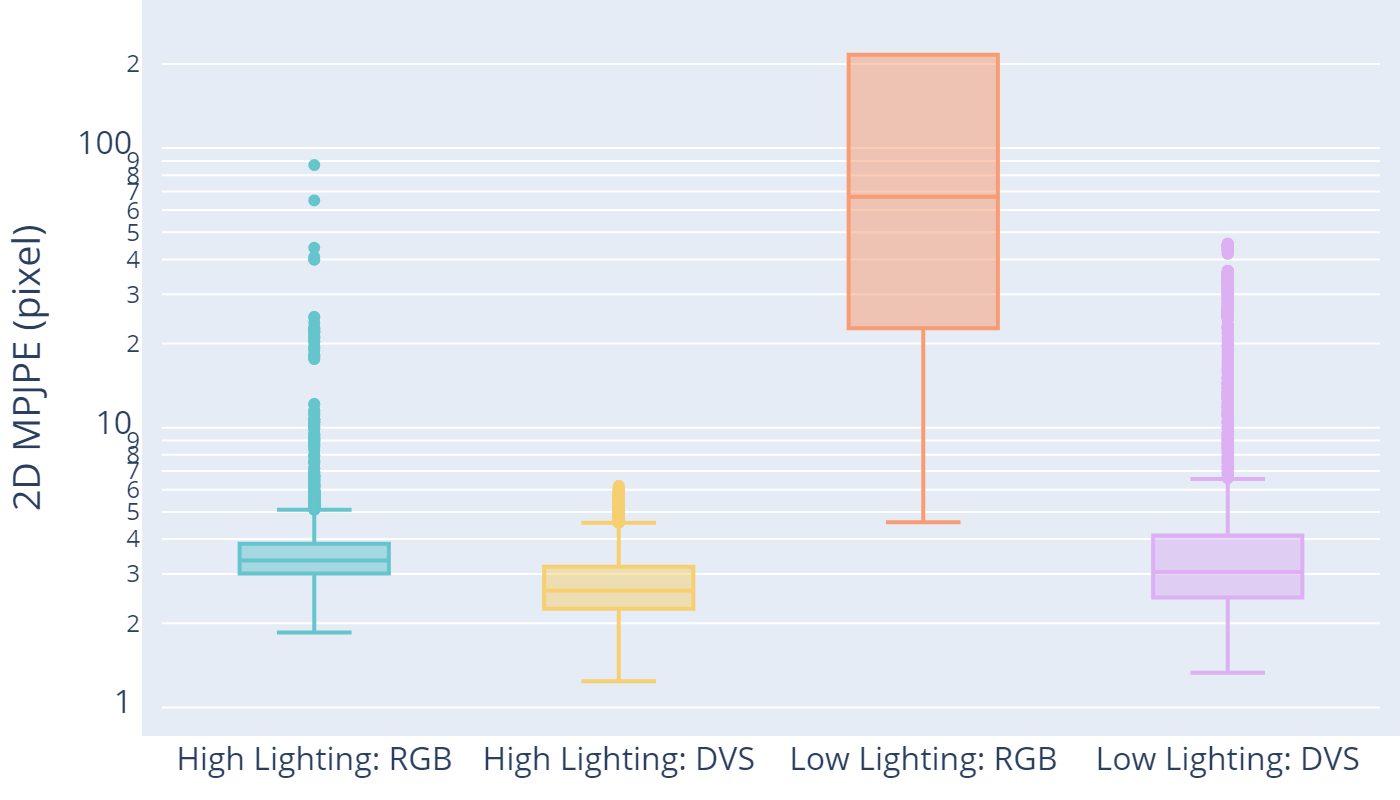}
    \caption{2D MPJPE Comparison between DVS and RGB.}
    \label{fig:box_plot}
\end{figure}

In order to show the superiority of DVS in low-lighting conditions, we also compare our result with RGB frame-based human pose estimation algorithm. We use the DVS camera DAVIS346, where both events and RGB frames are recorded synchronously. As the same device records both modalities via the same lens, there is no difference in the quantity of light captured by RGB and DVS. OpenPose \cite{openpose2019} proposed and implemented by Cao et al. is adopted as our RGB baseline. OpenPose is an accurate, fast, and robust 2D human pose estimation algorithm. Although there are differences in joint number and dimension, we calculated the 2D projection of 3D joints generated by \emph{YeLan}. We picked the closest 13 joints from all 25 OpenPose output joints to compare.
The comparison is conducted on paired DVS and RGB recordings from the dataset collected in the motion capture lab. Both high and low-lighting condition cases are included, and qualitative comparison is shown in Fig. \ref{fig:rgb_dvs_comp}. Furthermore, we made a box plot showing the 2D MPJPE of DVS and RGB in two lighting conditions in \ref{fig:box_plot}. From this figure, it is evident that \emph{YeLan} generates better and more stable predictions in all cases. Although the RGB-based method achieves good performance (marked by the low 2D MPJPE) in high-lighting conditions, the performance deteriorates significantly in the low lighting conditions. OpenPose fails to detect any human from frames due to motion blur and low SNR sensor data when the illumination is lower than a certain threshold. On the other hand, \emph{YeLan} shows strong robustness against lighting conditions changes and constantly generates high-quality predictions (marked by the low MPJPE in both cases). 

\subsection{RGB-Depth Camera}

\begin{table}[]
\caption{The comparison between RGB, RGBD, and event cameras. For FOV, the H, V, D stands for horizontal, vertical, diagonal FOV.}
\begin{tabular}{|l|l|l|l|}
\hline
Camera Type       & RGB (In DAVIS 346)                                                                              & RGBD (Intel RealSense 435i)                                                                                                                                                                                           & Event Camera (DAVIS346)                                                                         \\ \hline
Update Rate       & 40 Frames / Second                                                                              & 30 Frames / Second                                                                                                                                                                                                    & \textless 12 MEvents / second                                                                   \\ \hline
FOV               & \begin{tabular}[c]{@{}l@{}}Any large,\\ Dependent on lens.\\ Lens are replaceable.\end{tabular} & \begin{tabular}[c]{ll}IR Projector:&  Depth FOV:\\ H: 90±3&    H: 87\\ V: 63±3&     V: 58\\ D: 99±3&     D: 95\\ All FOVs are fixed.&\end{tabular}                                                                   & \begin{tabular}[c]{@{}l@{}}Any large,\\ Dependent on lens.\\ Lens are replaceable.\end{tabular} \\ \hline
Dynamic Range     & 56.7 dB                                                                                         & N/A                                                                                                                                                                                                                   & 120 dB                                                                                          \\ \hline
Power Consumption & 140mW                                                                                           & \begin{tabular}[c]{@{}l@{}}Maximum Power: 2850mW \\ (Measured on Windows 10) \end{tabular} & \begin{tabular}[c]{@{}l@{}}10-30mW \\ (activity dependent)\end{tabular}                         \\ \hline
Working Distance  & N/A                                                                                             & \begin{tabular}[c]{@{}l@{}}0.2$\sim$3m, varies with lighting\\ conditions, more limited for\\ HPE usage.\end{tabular}                                                                                                 & N/A                                                                                             \\ \hline
\end{tabular}
\end{table}


Besides the RGB camera, the depth camera is also widely used in digital dancing games. \cite{yun2014development, kamel2012xbox} These cameras are usually paired with RGB cameras, which grant them information from both domains. Compared to RGB cameras, these cameras have a better understanding of the 3D space, which results in a more accurate human segmentation and joint location estimation. RGBD camera-based human pose estimation is a well-established problem with many commercial products and pipelines \cite{rallis2018embodied, kitsikidis2014dance, alexiadis2011evaluating}, like the Kinect from Microsoft and the RealSense from Intel.

However, the depth camera also has several disadvantages. Depth cameras actively emit and recapture the reflected infrared to build the 3D point cloud. This procedure is significantly more power-hungry (about 200 times higher power consumption than the event camera) and suffers from many limitations. Firstly, due to the manufacturing and power consumption consideration, depth cameras' field of view (FOV) is usually fixed and small. Small FOV limits the sensing space coverage and restricts the dancer movement in a smaller space. Secondly, the distance between the target and the depth camera strongly impacts the detection accuracy. If the sensor is too close to the target (smaller than 0.5m), the two IR receivers have overlapped IR patterns which saturate the IR camera and result in an estimation failure \cite{ijgi6110349}. Moreover, when the distance is far, the detection rate drops drastically after a certain point, as the IR receiver cannot receive enough reflected IR light. The first two points together limit the functional dancing area and cause a substantial restriction on relative distance.

\begin{figure*}[h]
    \centering
    \includegraphics[width=0.6\linewidth]{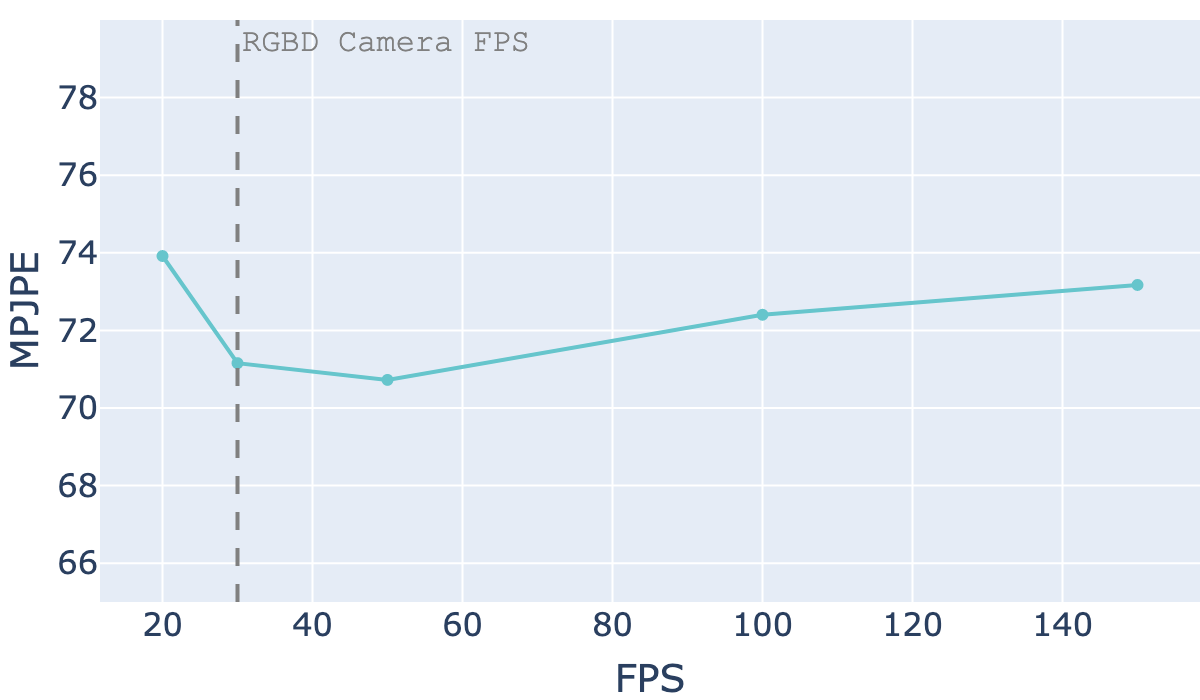}
    \caption{Stable human pose estimation of event camera on different FPS settings. The model used is only trained on 50FPS setting. On the contrary, the depth camera has a fixed 30FPS frame rate.}
    \label{fig:fps_var}
\end{figure*}

Thirdly, since the camera emits light itself, it could also be interfered with by other light sources, which is especially true for strong light sources like stage lights and solar lights. This characteristic restricts the application of depth cameras in outdoor and indoor spaces near the light source. 

Fourthly, compared with the event camera, the maximum RGBD camera FPS is usually very low, making it hard to be applied to track fast-paced dances or generate high-fidelity 3D digital dances. The two most commonly used RGBD cameras, i.e., Microsoft Azure Kinect and Intel RealSense, support at most 30 FPS when capturing full RGBD streams. On the contrary, event cameras can easily receive ten-millions level events per second, which gives them overwhelmingly huge advantages over RGBD cameras. We tested our model by generating representations from 20FPS to 150FPS (due to the restriction of ground truth label rate) on all the event streams from test subject eight. The result shows that the estimation accuracy is pretty stable on various frame rate settings (Fig. \ref{fig:fps_var}). 

\begin{figure*}[h]
    \centering
    \includegraphics[width=\linewidth]{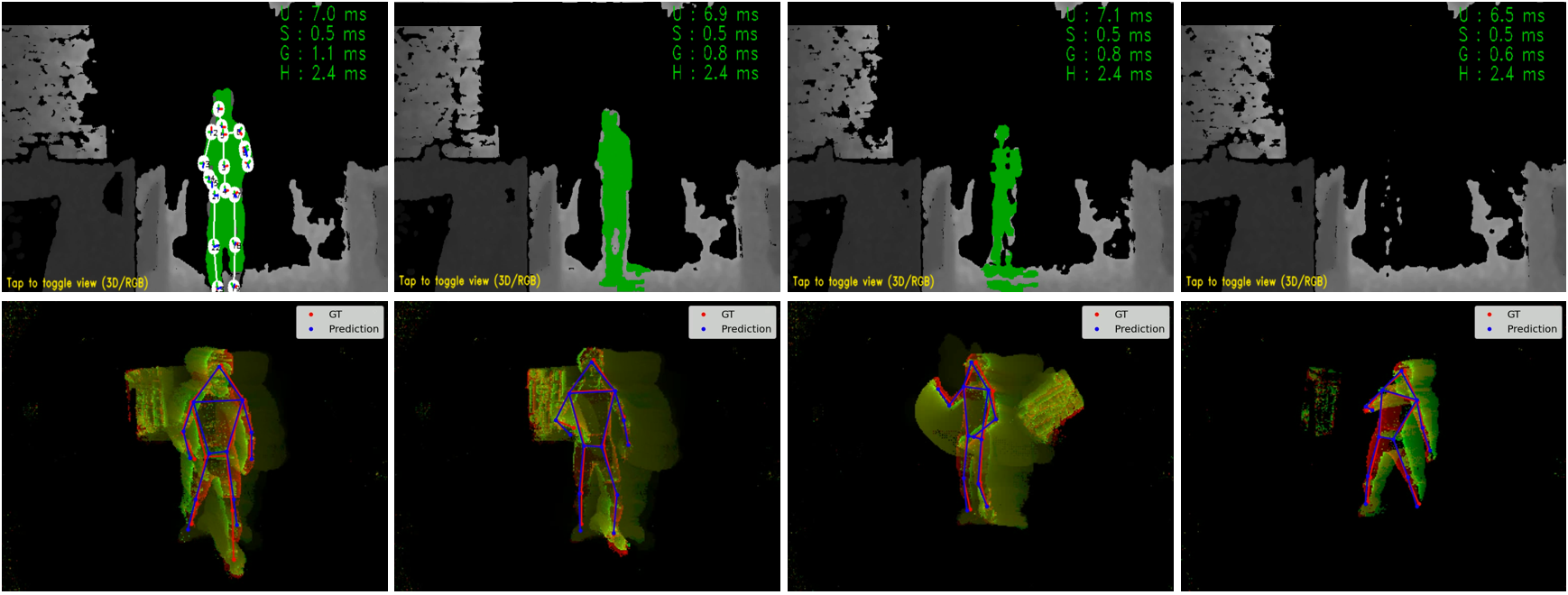}
    \caption{Qualitative comparison between RGB-Depth camera and YeLan in term of the distance between camera and human. The first row is the results from depth camera, while the second row is from YeLan. The distance rises from left to right.}
    \label{fig:depth_comp_dist}
\end{figure*}

Although we wanted to quantitatively compare RGBD and event camera for human pose estimation, we found out that these two types of cameras can not work together. When the depth camera is turned on, its built-in IR projector emits a grid of IR rays 30 times per second. In the eye of the event camera, the IR projection of the depth camera is visible and everything in the surrounding is constantly flickering at a high frequency with meshed dots, which negatively impacts the performance. Therefore, we made a qualitative comparison by separately recording two sessions of human dance by Intel RealSense 435i and DAVIS 346. During these two sessions, the participant moves from 2.5m to 3.5m with some dynamic dance patterns, and the results are shown in Fig. \ref{fig:depth_comp_dist}. As the figure shows, as the distance between the camera and the human increases, the RGB-Depth camera gradually fails to capture valid human shapes. The human pose estimation program also stops generating valid predictions from a certain point. The RGB-Depth camera-based 3D human pose estimation we chose for comparison is Nuitrack \cite{NuitrackFullBody}, a commercial product designed specifically for RGBD camera-based HPE problems. Although RGB and depth channels are all used, here, for visualization convenience, only the depth channel is shown, and the estimated human masks are shown in green.

\section{Limitation and Future Works}

Overall, YeLan clearly highlights the promise of the 3D human pose estimation in the context of dancing in the presence of different confounding factors. However, the current study has several limitations and we plan to systematically address them in future works. In the current work, one assumption was that the event camera is stationary and attached to a fixed tripod. Consequently, YeLan framework is not readily transferable to scenarios where the event camera can move and is mounted on a car or a drone. The current version of YeLan has been primarily tested in a single-person scenario and does not support a multiplayer dancing game. By including a multiperson segmentation and masking, we plan to extend YeLan for multi-person scenario. We also aim to explore energy efficient implementation in a neuromorphic computing platform such as Intel Loihi 2 \cite{LoihiIntelWikiChip}.

\section{Acknowledgement}
We would like to thank the Institute of Applied Life Sciences and the College of Information and Computer Sciences at UMass Amherst for providing start-up funds and laboratory support. Moreover, we wish to show our appreciation to the HDSI department at UCSD for their startup funding. The work was also partially funded by the DARPA TAMI grant (Project ID HR00112190041) and the directorate for computer science and engineering of NSF (Award Number 2124282). Lastly, I wish to extend my special thanks to miHoYo Co., Ltd. for providing high-quality character models used in synthetic data generation.

\section{Conclusion}

This work discussed the existing 3D HPE techniques applied in dance games, their strength, and limitations. We proposed an alternative and novel event camera-based method to tackle these shortcomings. We collected a real-world dance dataset by human subject studies and built a comprehensive motion-to-event simulator to generate massive fully-controllable, customizable, and labeled synthetic dance data to help pre-train the model. YeLan outperforms all baseline models in various challenging scenarios on both datasets. Lastly, we also did a thorough analysis as well as a comparison between different modalities, which clearly shows the superiority of YeLan in many aspects.



\bibliographystyle{ACM-Reference-Format}
\bibliography{ref}

\end{document}